%% file: main.tex
\documentclass[10pt,twocolumn,letterpaper]{article}

\usepackage[pagenumbers]{cvpr} 


\definecolor{cvprblue}{rgb}{0.21,0.49,0.74}
\usepackage[pagebackref,breaklinks,colorlinks,allcolors=cvprblue]{hyperref}

\def\paperID{9187} 
\def\confName{CVPR}
\def\confYear{2025}

\newcommand{\model}{{OmniGen}\xspace}
\usepackage{booktabs}       
\usepackage{amsfonts}       
\usepackage{nicefrac}       
\usepackage{microtype}      
\usepackage{xcolor}         
\usepackage{amsmath} 
\usepackage{natbib}
\usepackage{graphicx}

\title{OmniGen: Unified Image Generation}

\author{
 Shitao Xiao\thanks{Co-first authors}~~, Yueze Wang$^{*}$, Junjie Zhou$^{*}$, Huaying Yuan$^{*}$, \\
  Xingrun Xing, Ruiran Yan, Chaofan Li, Shuting Wang, Tiejun Huang, Zheng Liu\thanks{Corresponding authors} \\
 Beijing Academy of Artificial Intelligence \\
 \{stxiao, yzwang\}@baai.ac.cn, zhengliu1026@gmail.com
}

\begin{document}
\maketitle
\input{sec/0_abstract}  
\vspace{-0.6cm}
\input{sec/1_intro}

\input{sec/2_method}
\input{sec/3_exp}

\input{sec/4_related_work}

\input{sec/5_conclusion}
{
    \small
    \bibliographystyle{ieeenat_fullname}
    \bibliography{main}
}

\input{sec/X_suppl}

\end{document}

%% file: sec/0_abstract.tex
\begin{abstract}
The emergence of Large Language Models (LLMs) has unified language generation tasks and revolutionized human-machine interaction. 
However, in the realm of image generation, a unified model capable of handling various tasks within a single framework remains largely unexplored.
In this work, we introduce \model, a new diffusion model for unified image generation.  
\model is characterized by the following features:
1) \textbf{Unification}: \model not only demonstrates text-to-image generation capabilities but also inherently supports various downstream tasks, such as image editing, subject-driven generation, and visual-conditional generation. 
2) \textbf{Simplicity}: The architecture of \model is highly simplified, eliminating the need for additional plugins. Moreover, 
compared to existing diffusion models, it is more user-friendly and can complete complex tasks end-to-end through instructions without the need for extra intermediate steps, greatly simplifying the image generation workflow.
3) \textbf{Knowledge Transfer}: Benefit from learning in a unified format, \model effectively transfers knowledge across different tasks, manages unseen tasks and domains, and exhibits novel capabilities. 
We also explore the model's reasoning capabilities and potential applications of the chain-of-thought mechanism.
This work represents the first attempt at a general-purpose image generation model, and we will release our resources at \url{https://github.com/VectorSpaceLab/OmniGen} to foster future advancements.
\end{abstract}

%% file: sec/1_intro.tex
\section{Introduction}
\label{sec:intro}

\begin{figure}[t]
    \centering
    \includegraphics[width=\linewidth]{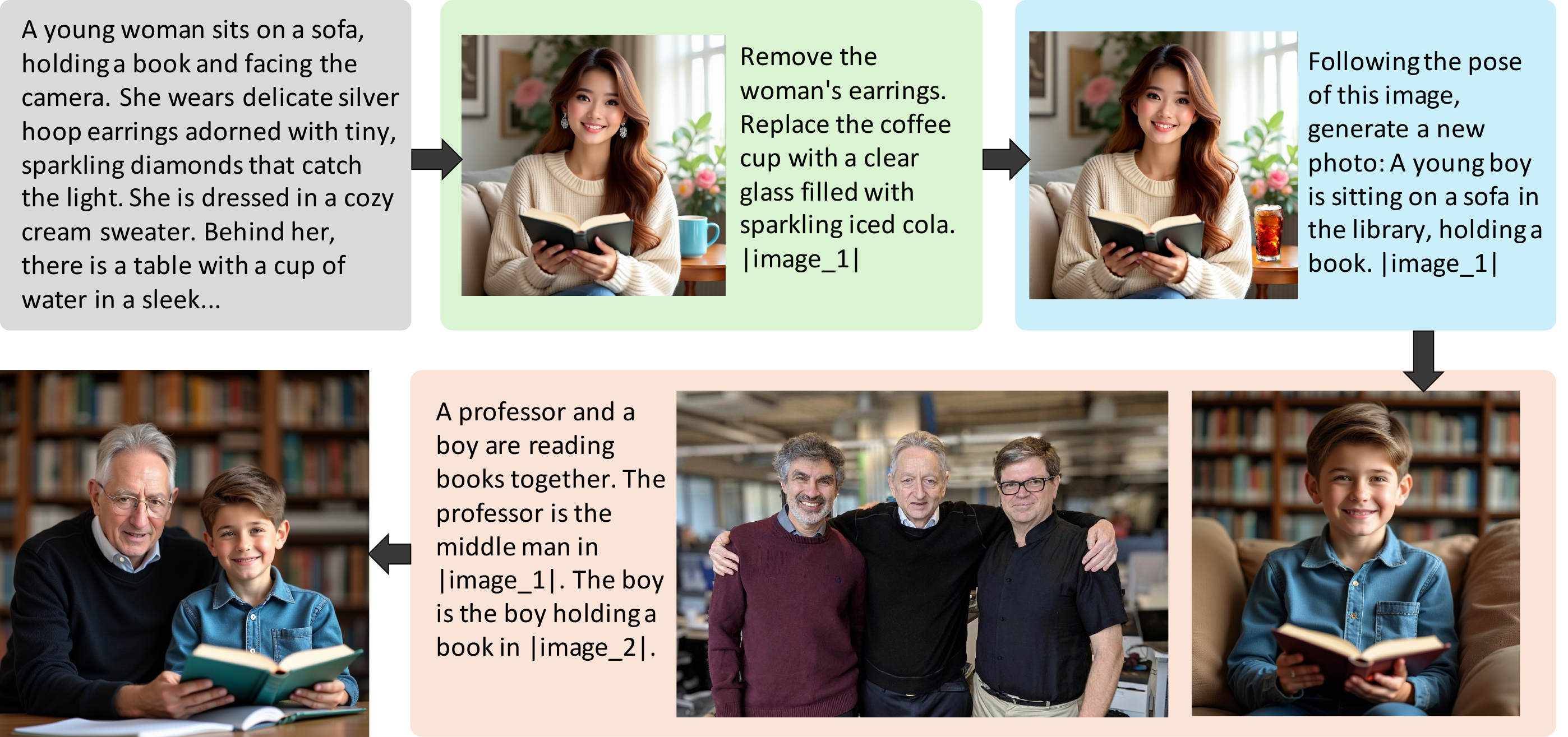}
    \vspace{-0.6cm}
    \caption{\model can flexibly follow instructions to complete various tasks, without the need for any additional plugins.
    For complex tasks (e.g., examples in the second line), it can also be completed end-to-end without cumbersome intermediate steps.}
    \label{fig:demo}
    \vspace{-0.6cm}
\end{figure}

The pursuit of Artificial General Intelligence (AGI) has intensified the demand for generative foundation models capable of handling various tasks within a single framework. In the field of Natural Language Processing (NLP), Large Language Models (LLMs) have become exemplary in achieving this goal, demonstrating remarkable versatility across numerous language tasks.
However, the realm of visual generation has yet to reveal a counterpart that matches the universality of LLMs. Current image generation models have demonstrated proficiency in specialized tasks. For instance, in the text-to-image generation filed,  models such as the SD series~\citep{LDM,podell2023sdxl,sd3}, DALL-E~\citep{dalle2}, and Imagen~\citep{imagen3} have made significant strides. Meanwhile, lots of efforts have been made to extend the capabilities of diffusion models for specific tasks, such as ControlNet~\citep{controlnet}, InstandID~\citep{wang2024instantid}, and InstructPix2Pix~\citep{brooks2023instructpix2pix}.
Currently, for each new task, designing a specific module and fine-tuning it is necessary, which hinders the development of image generation. What's worse, these task-specific models lead to a significant issue: we cannot accomplish tasks simply through input instructions but require a complex and cumbersome workflow involving multiple models and numerous steps.
For example, ControlNet needs to use a detector to extract conditions (such as pose estimation maps) and then loads the corresponding condition module. InstantID can only process single-person images and requires a face detection model to detect faces beforehand, followed by using a face encoder to encode the face. If users want to edit the generated images, additional models like InstructPix2Pix need to be loaded.



\begin{figure*}[ht]
    \centering
    \includegraphics[width=\linewidth]{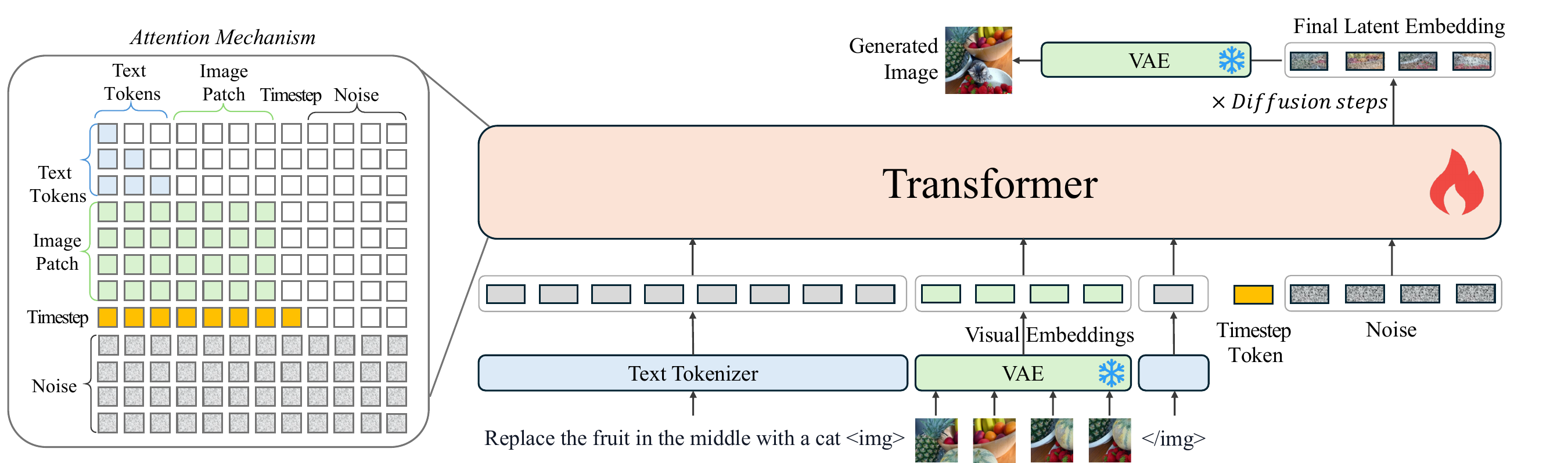}
    \vspace{-0.7cm}
    \caption{The framework of \model. Texts are tokenized into tokens, while input images are transformed into embedding via VAE. \model can accept free-form multi-modal prompts and generate images through the rectified flow approach.}
    \label{fig:model}
    \vspace{-0.5cm}
\end{figure*}

\textbf{Can a single model complete various tasks end-to-end through user instructions, similar to how ChatGPT handles language tasks?} We envision a future where image generation is made simple and flexible: that any tasks can be accomplished directly through user instructions. Motivated by this goal, we propose a unified framework: \model. As shown in Figure~\ref{fig:demo}, this framework allows for convenient and flexible image generation for any purposes, where no additional plugins or operations are needed. Given the flexibility in following arbitrary instructions, the new framework also help to inspire more interesting image generation tasks. 



Unlike popular diffusion models, \model features a very concise structure, comprising only two main components: a VAE and a transformer model. \model supports arbitrarily interleaved text and image inputs as conditions to guide image generation, instead of only accepting pure text or image. To train this architecture as a unified model, we construct a large-scale multi-task image generation dataset X2I, which unifies different tasks with one uniform format. We evaluate the well-trained model based on multiple benchmarks, demonstrating its superior generation capability compared to existing models. 
Remarkably, \model enables effective knowledge transfer across different scenarios, 
allowing it to handle unseen tasks and domains while also fostering the emergence of new abilities. 


Our contributions are summarized as follows: 
\begin{itemize}
    \item 
    We introduce \model, a unified image generation model that excels in multiple domains. The model demonstrates competitive text-to-image generation capabilities and inherently supports a variety of downstream tasks, such as controllable image generation and classic computer vision tasks. Furthermore, it can handle complex tasks end-to-end without any lengthy intermediate steps. To our knowledge, \model is the first image generation model to achieve such comprehensive functionality.
    \item We construct the first comprehensive image generation dataset named X2I, which stands for "anything to image". This dataset covers a wide range of image generation tasks, all of which are standardized into a unified format.
    \item By unified training on the multi-task dataset, \model can apply learned knowledge to tackle unseen tasks and domains, as well as exhibit new capabilities. Additionally, we explore the model's reasoning capabilities and chain-of-thought mechanism.
    \item OmniGen takes an initial step toward a foundational model for general image generation. 
    We will open-source the relevant resources (model, code, and data) and hope this work can provide some insights for future image generation models.
\end{itemize}

%% file: sec/2_method.tex
\section{\model}
\label{sec:model}

In this section, we present the details of \model framework, including the model architecture and training method.

\subsection{Model Design}
\label{sec:design}

\textbf{Network Architecture}.
The design principles of \model are universality and conciseness.
As illustrated in Figure~\ref{fig:model}, \model adopts an architecture comprised of a Variational Autoencoder (VAE)~\citep{vae} and a pre-trained large transformer model.
Specifically, we use the VAE from SDXL~\citep{podell2023sdxl} to extract continuous visual features from images. 
The transformer model initialized by Phi-3~\citep{abdin2024phi3} generates images based on instructions that serve as conditions. Only VAE is frozen during training.
Unlike state-of-the-art diffusion models that require additional encoders to pre-process conditional information (such as CLIP text encoder and image encoder), \model inherently encodes conditional inputs by itself, significantly simplifying the pipeline. Furthermore, \model jointly models text and images within a single model, rather than independently modeling different input conditions with separate encoders as in existing works~\citep{wei2023elite, xiao2023fastcomposer, ye2023ipadapter, wang2024instantid, chen2024subject} which lacks interaction between different modality conditions.

\textbf{Input Representation}.
The input to the model can be multi-modal interleaved text and images in free form. 
We utilize the tokenizer of Phi-3 to process text.
For images, we firstly employ a VAE with a simple linear layer to extract latent representations. 
Then, they are flattened into a sequence of visual tokens by linearly embedding each patch in the latent space. 
Following~\citep{dit}, the patch size is set to 2.
We apply standard frequency-based positional embeddings to visual tokens, and process images with varying aspect ratios as SD3~\citep{sd3}. 
In addition, we encapsulate each image sequence with two special tokens: ``$<$img$>$'' and ``$<$/img$>$'' before inserting it into the sequence of text tokens. 
We also append the timestep embedding~\citep{dit} at the end of the input sequence.
We do not need any task-specific special tokens to indicate the task type.

\textbf{Attention Mechanism}.
Different from text, which can be decomposed into discrete tokens to model, we argue that the image should be modeled as a whole based on its nature. 
Therefore, we modify the common causal attention mechanism in LLM by integrating it with the bidirectional attention as illustrated in Figure~\ref{fig:model}.
Specifically, we apply causal attention to each element in the sequence, but apply bidirectional attention within each image sequence. 
This allows each patch to pay attention to other patches within the same image, while ensuring that each image can only attend to other images or text sequences that have appeared previously.

\textbf{Inference}.
During inference, we randomly sample a Gaussian noise and then apply the flow matching method to predict the target velocity, iterating multiple steps to obtain the final latent representation.  
Finally, we use a VAE decoder to decode the latent representation into the predicted image. 
Thanks to the attention mechanism, \model can accelerate inference like LLMs by using kv-cache: storing previous and current key and value states of the input conditions on the GPU to compute attention without redundant computations.

\subsection{Training Strategy}
\label{sec:train}

\textbf{Train objective}.
In this work, we use rectified flow~\citep{liu2022flow} to optimize the parameters of model. 
Different from the DDPM~\citep{ddpm}, flow matching conducts the forward process by linearly interpolating between noise and data in a straight line.
At the step $t$, $\mathbf{x}_t$ is defined as: $\mathbf{x}_t = t \mathbf{x} + (1-t) \boldsymbol{\epsilon}$,
where $\mathbf{x}$ is the original data, and $\boldsymbol{\epsilon} \sim \mathcal{N}(0, 1)$ is the Gaussian noise. 
The model is trained to directly regress
the target velocity given the noised data $\mathbf{x}_t$, timestep $t$, and condition information $c$. Specifically, the objective is to minimize the mean squared error loss:
\begin{equation}
\mathcal{L} = \mathbb{E}_{}\big[||( \mathbf{x} - \boldsymbol{\epsilon}) - v_{\theta}\big (\mathbf{x}_t,~ t, ~ c \big)||^2 \big ].
\end{equation}

For image editing tasks, the objective is to modify specific regions of the input image while keeping other areas unchanged. Therefore, the difference between the input image and the target image is often small, which allows the model to learn an unexpected shortcut: simply copying the input image as the output to make the related training loss very low. To mitigate this phenomenon, we amplify the loss in the regions of the image where changes occur. More specifically, we calculate the loss weights for each region based on latent representations of input image $\mathbf{x'}$ and target image $\mathbf{x}$: 
\begin{equation}
   \quad w_{i,j} = 
    \begin{cases} 
        1 & \text{if } \mathbf{x}_{i,j} = \mathbf{x'}_{i,j} \\
        \frac{1}{||\mathbf{x} - \mathbf{x'}||^2} & \text{if } \mathbf{x}_{i,j} \neq \mathbf{x'}_{i,j} 
    \end{cases}
\end{equation}
Consequently, regions with alterations are assigned significantly higher weights than those without changes, guiding the model to focus on the areas to be modified.

\textbf{Training Pipeline}.
Following previous work~\citep{sd3,gao2024lumina,chen2023pixart}, we gradually increase the image resolution during the training process. Low resolution is data-efficient, while high resolution can enhance the aesthetic quality of the generated images. Detailed information for each training stage is presented in Table~\ref{tab:pipeline}. We adopt the AdamW~\citep{adamw} with $\beta = (0.9, 0.999)$ as the optimizer.
All experiments are conducted on 104 A800 GPUs.

\begin{table}[t]
    \centering
    \resizebox{1.\linewidth}{!}{ 
    \begin{tabular}{ccccc}
    \toprule
    Stage & Resolution &  Steps (K) & Batch Size & Learning Rate \\
    \midrule
    1 & 256$\times$256 &   500 & 1040 & 1e-4 \\
    2 & 512$\times$512 &   300 & 520 & 1e-4 \\
    3 & 1024$\times$1024   & 100 & 208 &  4e-5 \\
    4 & 2240$\times$2240 & 30 & 104 & 2e-5 \\
    5 & Multiple &  80 & 104 & 2e-5 \\
    \bottomrule
    \end{tabular}
    }
    \vspace{-0.3cm}
    \caption{Details about each training stage of \model.}
    \label{tab:pipeline}
    \vspace{-0.7cm}
\end{table}

\section{X2I Dataset}

\begin{figure}[t]
    \centering
    \includegraphics[width=\linewidth]{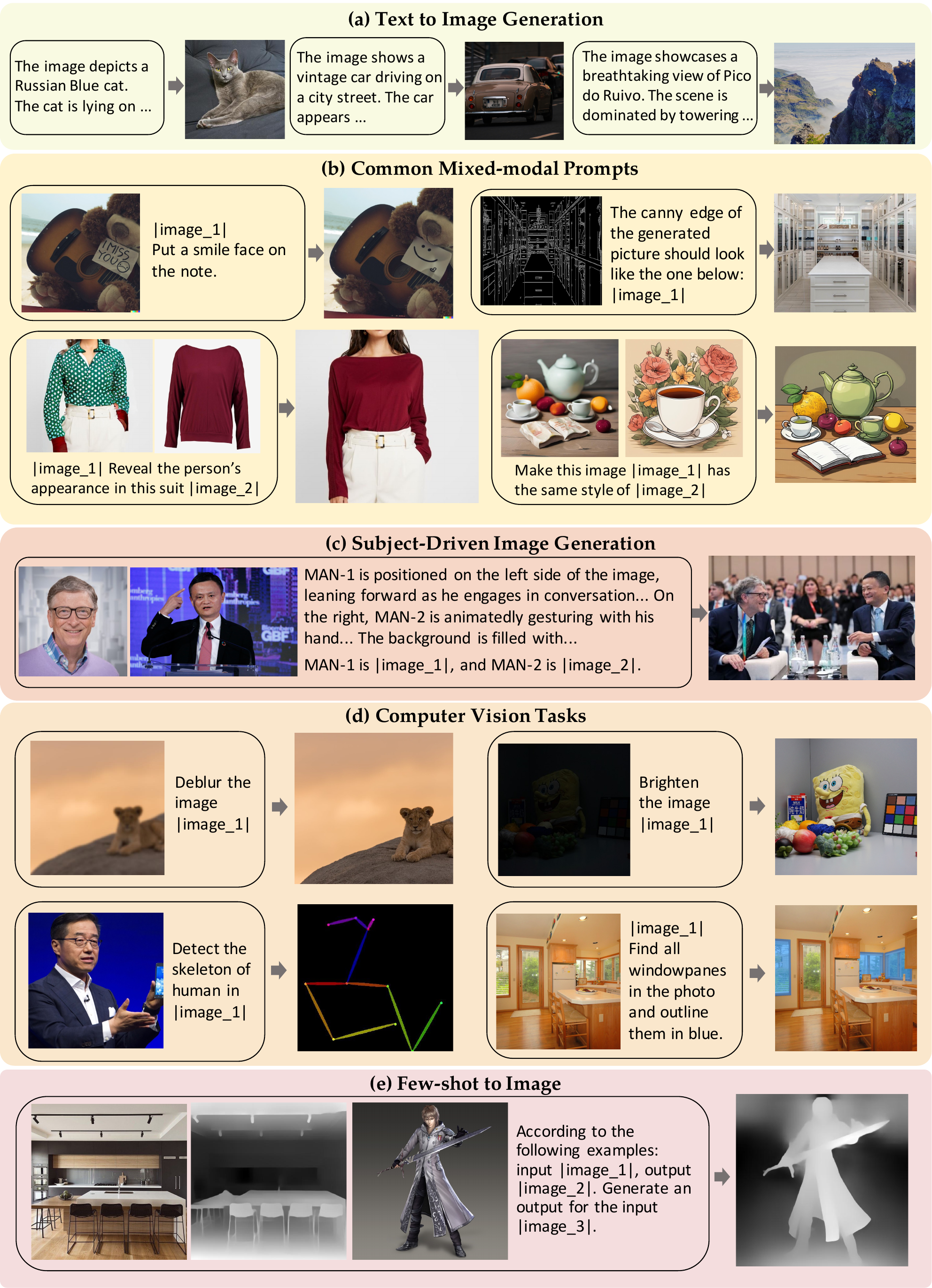}
    \vspace{-0.5cm}
    \caption{Examples of X2I dataset. We standardized the input of all tasks into an interleaved image-text sequence format. The placeholder $|$image\_i$|$ represents the position of the $i$-th input image.}
    \label{fig:data-overview}
    \vspace{-0.6cm}
\end{figure}

To achieve robust multi-task processing capabilities, it is essential to train the model on large-scale and diverse datasets. 
However, in the field of image generation, a readily available large-scale and diverse dataset has yet to emerge. 
In this work, we have constructed a large-scale unified image generation dataset for the first time, which we refer to as the \textbf{X2I} dataset, meaning "anything to image". 
We have converted all data into a unified format, and Figure~\ref{fig:data-overview} presents some examples of the X2I dataset. 
The entire dataset comprises approximately 0.1 billion images. 
A detailed description of the composition will be provided in the following sections.

\subsection{Text to Image}

The input for this subset of data is plain text.
We collect multiple open-source datasets from various sources: Recap-DataComp~\citep{li2024recaption}(a subset of 56M images), SAM-LLaVA~\citep{chen2023pixart}, ShareGPT4V~\citep{chen2023sharegpt4v}, LAION-Aesthetic~\citep{schuhmann2022laion}(a subset of 4M images), ALLaVA-4V~\citep{chen2024allava}, DOCCI~\citep{onoe2024docci}, DenseFusion~\citep{li2024DenseFusion} and JourneyDB~\citep{sun2024journeydb}. 
While these datasets are large in quantity, their image quality is not always high enough. 
In the early stages of training, we use them to learn a broad range of image-text matching relationships and diverse knowledge. 
After stage 3, we utilize an internal collection of 16M high-quality images to enhance the aesthetic quality of the generated images. 
We use the InternVL2~\citep{internvlm2} to create synthetic annotations for internal data and LAION-Aesthetic.

\subsection{Multi-modal to Image}

Different from existing diffusion models, \model can accept more flexible multi-modal instruction, thus adapting to a variety of data and task types.

\subsubsection{Common Mixed-modal Prompts}

The input of this portion of data is arbitrarily interleaved text and images. 
We collect the data from multiple tasks and sources: image editing (MagicBrush~\citep{zhang2024magicbrush}, InstructPix2Pix~\citep{brooks2023instructpix2pix} and SEED-edit~\citep{ge2024seededit}, ), human motion (Something-Something~\citep{goyal2017something}), virtual try-on (HR-VITON~\citep{lee2022hrviton} and FashionTryon~\citep{fashiontryon}), and style transfer (StyleBooth~\citep{han2024stylebooth}). 
The issue of utilizing additional visual conditions for finer-grained spatial control has garnered widespread attention\citep{controlnet, controlnetplus}. 
We employ the MultiGen~\citep{qin2023unicontrol} dataset to learn this function, and select six representative visual conditions: Canny, HED, Depth, Skeleton, Bounding Box, and Segmentation. 
These types of tasks take text prompts with specific visual conditions (such as segmentation maps, and human pose maps) as multi-modal inputs. 
We standardize all tasks into the input-output pair format as shown in Figure~\ref{fig:data-overview}-(b).

\subsubsection{Subject-driven Image Generation}

We construct both a large-scale foundational dataset (GRIT-Entity dataset) and a high-quality advanced dataset (Web Images dataset) for subject-driven image generation. 
For the GRIT-Entity dataset, we leverage the GRIT dataset~\citep{kosmos-2}, which annotates object names within images. 
Using these annotations, we apply GroundingDINO~\citep{groundingdino} for open-set text-to-bounding-box detection. 
Based on the grounded bounding boxes, we employ SAM~\citep{sam} to segment object masks from cropped images. 
We further use the MS-Diffusion~\citep{ms-diffusion} to repaint the object images for higher quality and diversity of data. 
The process of data construction and the final instruction format are illustrated in supplementary material. 
Through these methods, we acquire 6M data pairs.

Although the GRIT-based approach provides a substantial amount of samples, the input data extracted directly from original images will tend to lead the model to fall into a simple copy-and-paste pattern. 
To fully unleash the subject-driven image generation capability of \model, we construct a high-quality natural images dataset from the web. 
We use the search engine to collect multiple different images of the same person, using one as the input and another as the output. 
An example is shown in Figure~\ref{fig:data-overview}-(c).
More details about the construction process are given in the supplementary material.

\subsubsection{Computer Vision Tasks}

We introduce classic computer vision tasks to expand the boundaries of the model's generative capabilities.
For low-level vision tasks (low-light image enhancement~\citep{lightenhance}, deraining~\citep{derain}, deblurring~\citep{gopro}, inpainting~\citep{qin2023unicontrol}, outpainting~\citep{qin2023unicontrol} and colorization~\citep{schuhmann2022laion}), where the annotation itself is an image, we only add text instructions, which are randomly sampled from instructions generated by GPT-4o. 
For high-level vision tasks, we choose to represent all annotations as images. 
We use images from LAION~\citep{schuhmann2022laion} as source images and annotations from~\citep{qin2023unicontrol} as targets to construct image pairs (such as source image and its human pose mapping). 
The annotations include human pose, depth mapping, canny, Hed edge, and segmentation.
Additionally, we also use several datasets for referring image segmentation, including RefCOCO~\citep{refcoco}, ADE20k~\citep{ade20k}, and ReasonSeg~\citep{lai2023lisa}. 
As shown in Figure~\ref{fig:data-overview}-(d), the input is the source image and a natural language expression, the output is an image with the corresponding object highlighted in blue. 


\subsection{Few-shot to Image}
We build a few-shot to image dataset to stimulate the in-context learning ability of the model. 
Specifically, for each task described in the preceding sections, we randomly selected a few examples and combined the original input with these examples to form new inputs. 
The specific data format is shown in Figure~\ref{fig:data-overview}-(e). 
Due to limitations in training resources, we opted to use only one example to improve training efficiency.

%% file: sec/3_exp.tex
\section{Experimental Results}

The default inference step is set to 50. We use the classifier-free guidance for both text condition and image condition following ~\citep{brooks2023instructpix2pix}. 
The default guidance scale is set to 2.5, and the image guidance scale is set to 1.6. 
For image editing task, we increase the image guidance scale to 2.

\subsection{Image Generation}

\subsubsection{Qualitative Results}

\begin{figure}
    \centering
    \includegraphics[width=\linewidth]{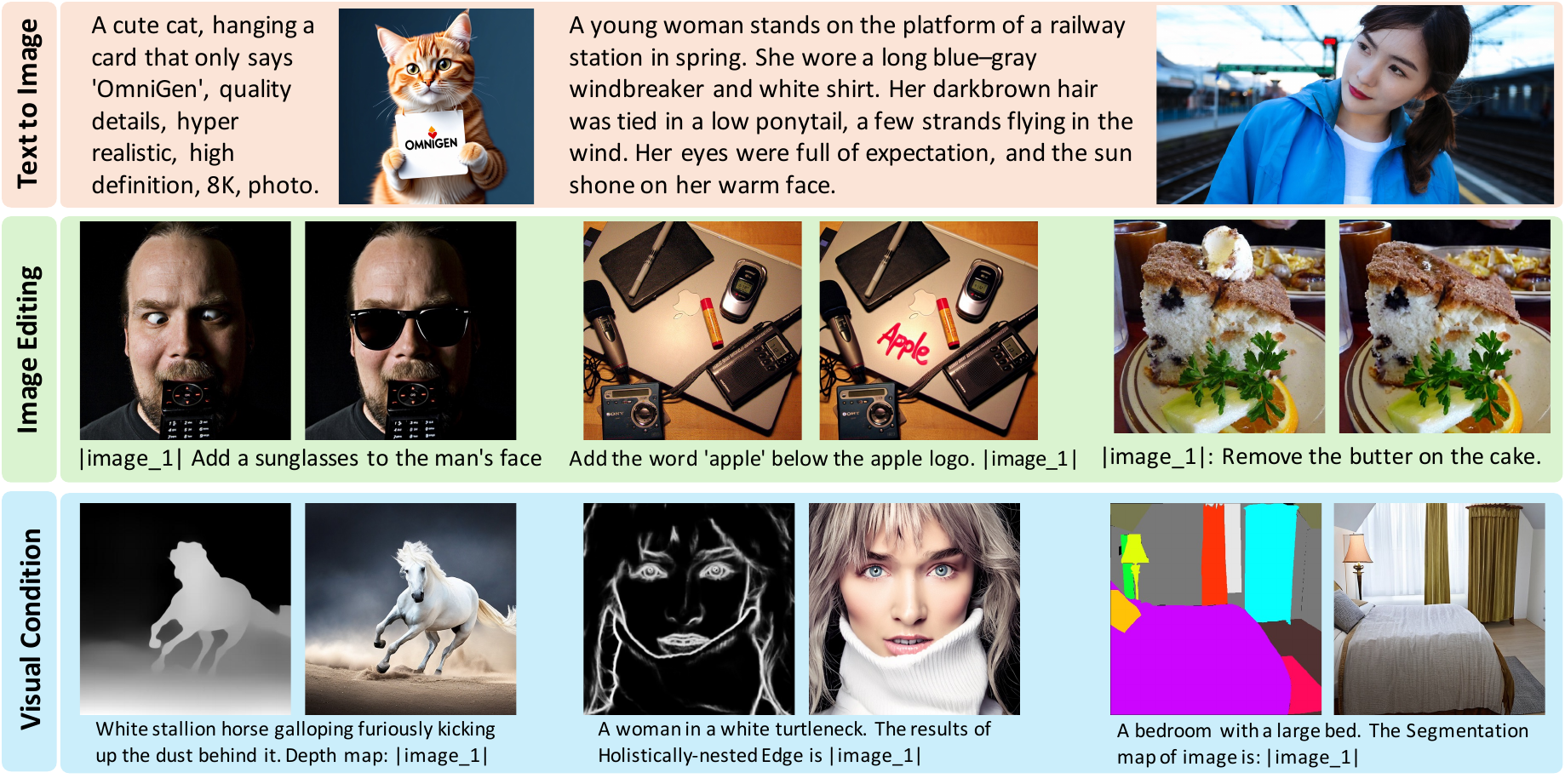}
    \vspace{-0.7cm}
    \caption{The results of different image generation tasks.}
    \label{fig:image_gen}
    \vspace{-0.3cm}
\end{figure}

\begin{figure}
    \centering
    \includegraphics[width=\linewidth]{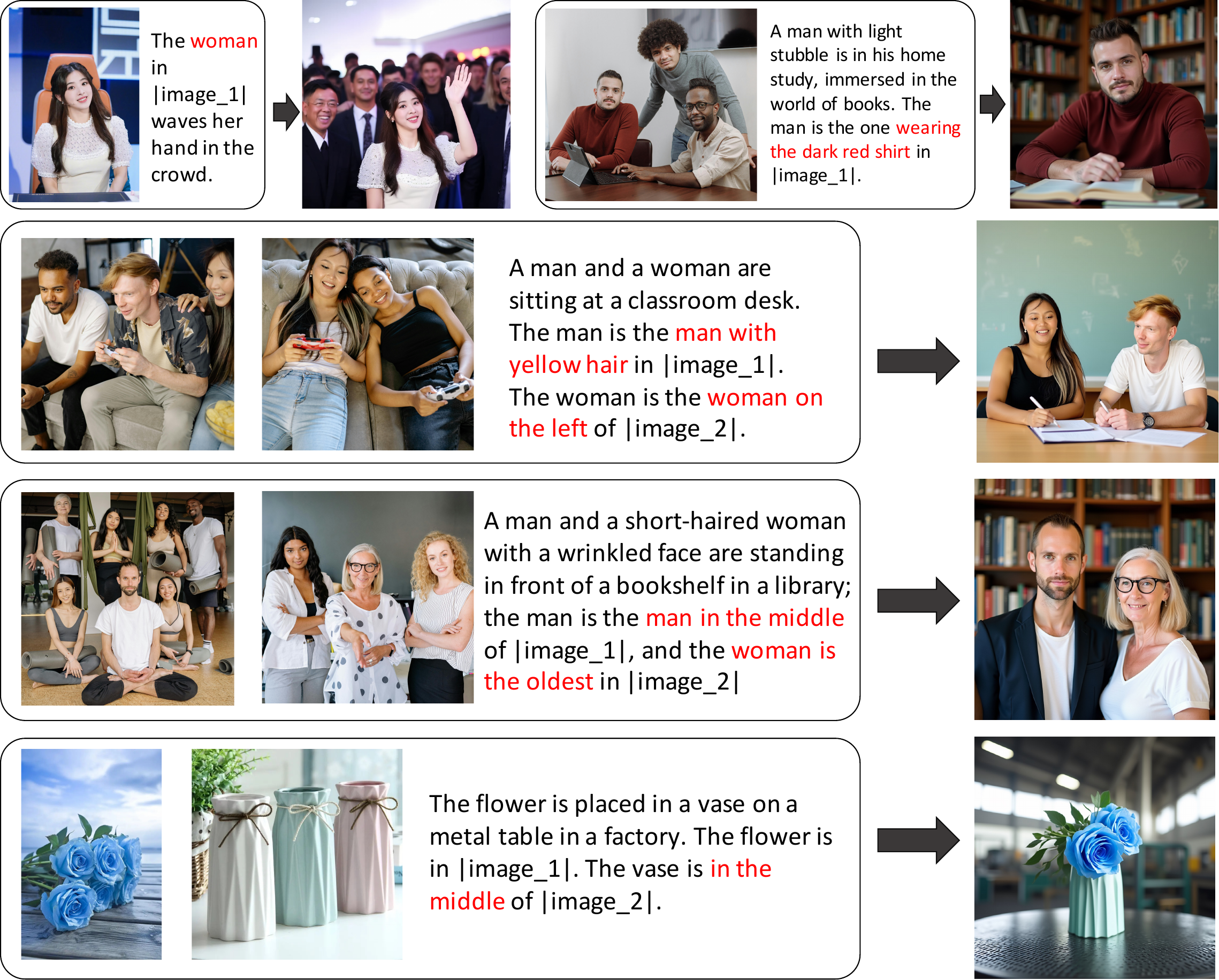}
    \vspace{-0.7cm}
    \caption{Subject-driven generation. When the reference image contains multiple objects, \model can automatically identify the required objects based on textual instructions without the need for additional preprocessing steps like image cropping or face recognition.}
    \label{fig:entity}
    \vspace{-0.6cm}
\end{figure}

Figure~\ref{fig:image_gen} summarizes the results of different image generation tasks, including text-to-image, image editing, and visual conditional. 
These results demonstrate that \model can handle various downstream tasks based on multi-modal instructions.

Figure~\ref{fig:entity} presents the outcomes of the subject-driven generation task. 
\model can extract the required objects from the given reference images and generate new images accordingly. Furthermore, when the reference image contains multiple objects, the model can directly select the needed objects based on textual instructions (e.g., the one wearing
the dark red shirt in image) without requiring additional preprocessing steps. 
In contrast, existing models~\citep{wang2024instantid,guo2024pulid} require manually cropping the target person from a multi-person image, detecting faces with a face detector, encoding faces with a face encoder, and inputting them into the diffusion model through a cross-attention module. Without these complex processes, OmniGen can complete the task in a single end-to-end step.

More qualitative results are provided in supplementary materials.

\subsubsection{Text to Image}
Following~\citep{sd3}, we evaluate text-to-image generation capability on the GenEval~\citep{ghosh2024geneval} benchmark. We compare the performance of \model with the reported results of other popular image generation models, as summarized in Table~\ref{tab:geneval}. 
Surprisingly, \model achieves similar performance compared to the current diffusion models, such as SD3, which underscores the effectiveness of our framework.
Notably, \model has only 3.8 billion parameters, whereas the SD3 model has a total of 12.7 billion parameters. 
The architecture of \model is significantly simplified, eliminating the cost of an additional encoder, thereby greatly enhancing the efficiency of parameter utilization. 

\begin{table}[t]
    \centering
     \resizebox{1.\linewidth}{!}{
    \begin{tabular}{lcccccccc}
        \toprule
 &  & &  Single & Two &  &  &  & Attribute  \\
 \textbf{Model} & \textbf{Params}  & \textbf{Overall} & object & object & Counting & Colors & Position & binding  \\
\cmidrule(lr){1-1}
\cmidrule(lr){2-2}
\cmidrule(lr){3-3}
\cmidrule(lr){4-9}
SDv1.5 & 0.9B+0.1B$^*$  & 0.43  & 0.97 & 0.38 & 0.35 & 0.76 & 0.04 & 0.06  \\
SDv2.1 & 0.9B+0.4B$^*$  & 0.50  &  \underline{0.98} & 0.51 & 0.44 & \textbf{0.85} & 0.07 & 0.17  \\
SD-XL & 2.6B+0.8B$^*$  & 0.55 & \underline{0.98} & {0.74} & 0.39 & \textbf{0.85} & {0.15} & 0.23    \\
DALLE-2 & 3.5B+1.0B$^*$  & 0.52 & 0.94 & 0.66 & 0.49 & 0.77 & 0.10 & 0.19 \\
DALLE-3 & --  & 0.67 & 0.96 & \textbf{0.87} & 0.47 & \underline{0.83} & \textbf{0.43} & \underline{0.45} \\
IF-XL & 5.5B+4.7B$^*$  & {0.61} & {0.97} & {0.74} & \textbf{0.66} & 0.81 & {0.13} & {0.35}    \\
SD3  & 8.0B+4.7B$^*$  & \underline{0.68} & \underline{0.98} &  0.84 & \textbf{0.66} & 0.74 & \underline{0.40} & 0.43  \\
\model  & 3.8B  & \textbf{0.70}  & \textbf{0.99} & \underline{0.86} & \underline{0.64} & \textbf{0.85} & 0.31 & \textbf{0.55}  \\
        \bottomrule
    \end{tabular}}
    \vspace{-0.3cm}
    \caption{Results of GenEval Benchmark. $^*$ means the parameter of the frozen text encoder. Our model achieves comparable performance with a relatively small parameter scale.}
    \label{tab:geneval}
    \vspace{-0.5cm}
\end{table}

\subsubsection{Multi-modal to Image}

\begin{table}[h]
\small
\centering
\setlength\tabcolsep{1.5pt}
\begin{minipage}[!t]{0.47\linewidth}
    \centering
    \begin{tabular}	{l c c}
    \toprule
    \textbf{CLIP Simi} & \textbf{-T}$\uparrow$ & \textbf{-I}$\uparrow$   \\
    \midrule
    I-Pix2Pix~\citep{brooks2023instructpix2pix} & 0.219 & 0.834   \\
    MagicBrush~\citep{zhang2024magicbrush} & 0.222 & \underline{0.838}   \\
    PnP~\citep{pnp} & 0.089 & 0.521   \\
    Null-Text~\citep{mokady2023null} & \textbf{0.236} & 0.761   \\
    EMU-Edit~\citep{sheynin2024emuedit} & \underline{0.231} & \textbf{0.859}   \\
    \model & \underline{0.231} & 0.829     \\    
    \bottomrule
    \end{tabular}
\end{minipage}
\hfill
\begin{minipage}[!t]{0.47\linewidth}
    \centering
    \begin{tabular}	{l c c}
    \toprule
    \textbf{CLIP Simi} & \textbf{-T}$\uparrow$ & \textbf{-I}$\uparrow$   \\
    \midrule
    Tex-Inv~\citep{text_inv} & 0.255 & 0.780  \\
    DreamBooth~\citep{ruiz2023dreambooth}  & \underline{0.305}  & 0.803 \\
    ReImagen~\citep{chen2022re_imagen} & 0.270 & 0.740  \\
    SuTI~\citep{suti}  & {0.304}  & \underline{0.819} \\
    Kosmos-G~\citep{pan2023kosmosg}  & 0.287 & \textbf{0.847}  \\
    \model & \textbf{0.315} & {0.801}  \\
    \bottomrule
    \end{tabular}
\end{minipage}
\vspace{-0.3cm}
\caption{\small\textbf{Left}: Results on EMU-Edit test data. \textbf{Right}: Results on DreamBench. As a universal model, \model demonstrates performance comparable to that of the best proprietary models.
}
\label{tab:quantitative}
\vspace{-0.5cm}
\end{table}

\begin{table}[h]
\centering
\resizebox{1.\linewidth}{!}{ 
\begin{tabular}{l|c|c|c|c}
\toprule
& \textbf{Seg. Mask} & \textbf{Canny Edge} & \textbf{Hed Edge} & \textbf{Depth Map} \\
 & (mIoU$\uparrow$) & (F1 Score$\uparrow$) & (SSIM$\uparrow$) & (RMSE$\downarrow$) \\
\midrule
T2I-Adapter~\citep{t2iadapter}  & 12.61  & 23.65 & -  & 48.40 \\
Gligen~\citep{li2023gligen}  & 23.78  & 26.94 & 0.5634  & 38.83 \\
Uni-ControlNet~\citep{zhao2024uni}  & 19.39  & 27.32 & 0.6910 & 40.65 \\
UniControl~\citep{qin2023unicontrol}  & 25.44  & 30.82 & 0.7969 & 39.18 \\
ControlNet~\citep{controlnet}  & 32.55  & 34.65 & 0.7621 & 35.90 \\ 
ControlNet++~\citep{controlnetplus} & \textbf{43.64} & \underline{37.04} & \underline{0.8097} & \textbf{28.32} \\ 
\model & \underline{40.06} & \textbf{38.96} & \textbf{0.8332} & \underline{31.71} \\
\bottomrule
\end{tabular}
}
\vspace{-0.3cm}
\caption{Comparison with SOTA methods on controllable image generation. $\uparrow$ indicates higher result is better, while $\downarrow$ means lower is better. }
\label{tab:controllability}
\vspace{-0.5cm}
\end{table}

We evaluate the image editing on EMU-Edit~\citep{sheynin2024emuedit} dataset and subject-driven generation capability on DreamBench~\citep{ruiz2023dreambooth}.
We use CLIP-T to measure how well the model followed the instructions, while CLIP-I similarity scores measure the model's ability to preserve elements from the source image.
As shown in Table~\ref{tab:quantitative}, \model exhibits comparable performance to the current state-of-the-art models.  
Noted for subject-driven generation task, we select one image of the specific object as input instead of fine-tuning the model like DreamBooth.

In Table~\ref{tab:controllability}, we use the dataset and script from ~\citep{controlnetplus} to evaluate the generation capability based on visual conditions.
For each condition, the controllability is evaluated by measuring the similarity between the input conditions and the extracted conditions from generated images of diffusion models.
We can find that \model achieves competitive results for all conditions.


\begin{figure}
    \centering
    \includegraphics[width=\linewidth]{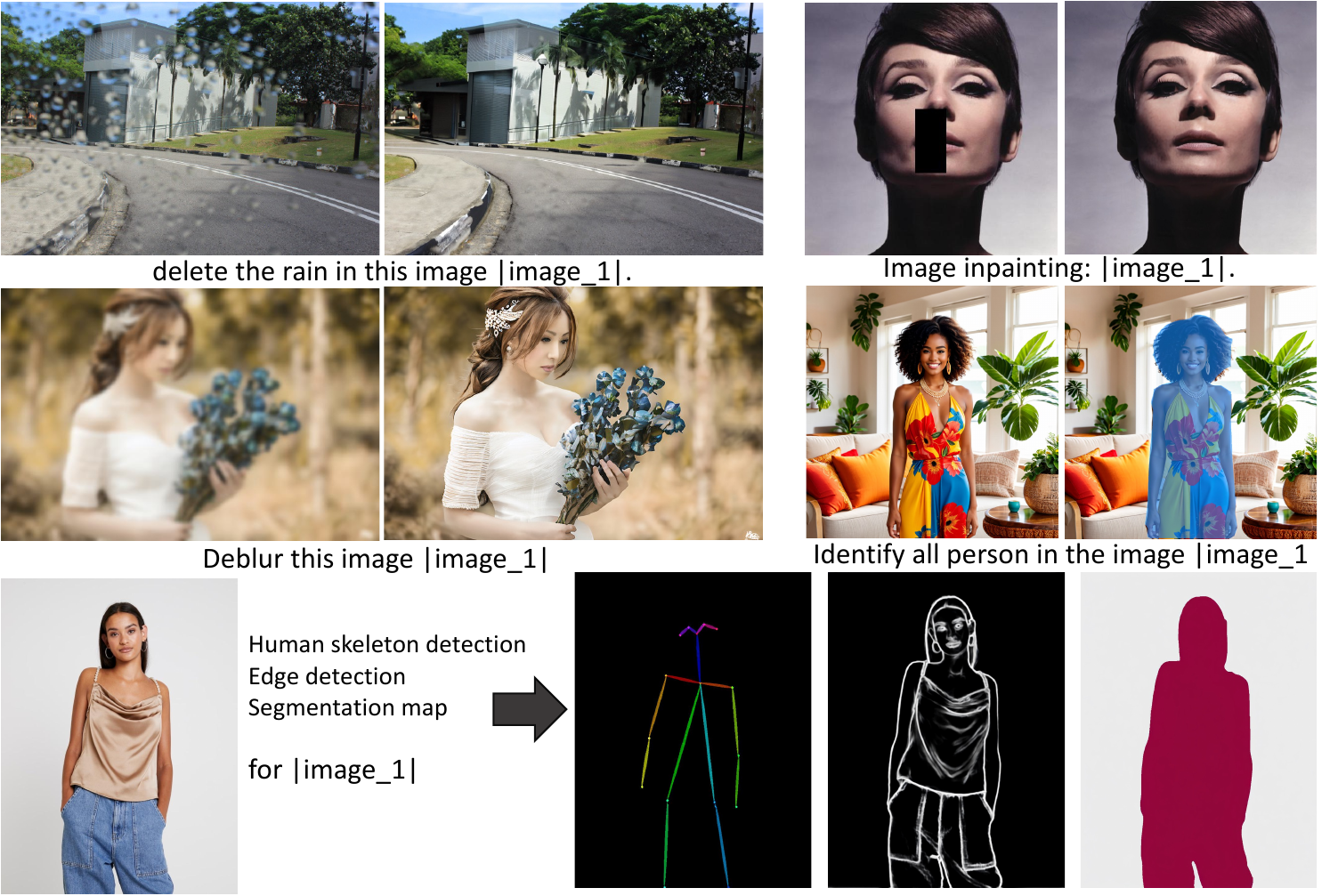}
    \vspace{-0.6cm}
    \caption{The results of \model in various traditional vision tasks.}
    \label{fig:cv}
    \vspace{-0.3cm}
\end{figure}

\begin{figure}
    \centering
    \includegraphics[width=\linewidth]{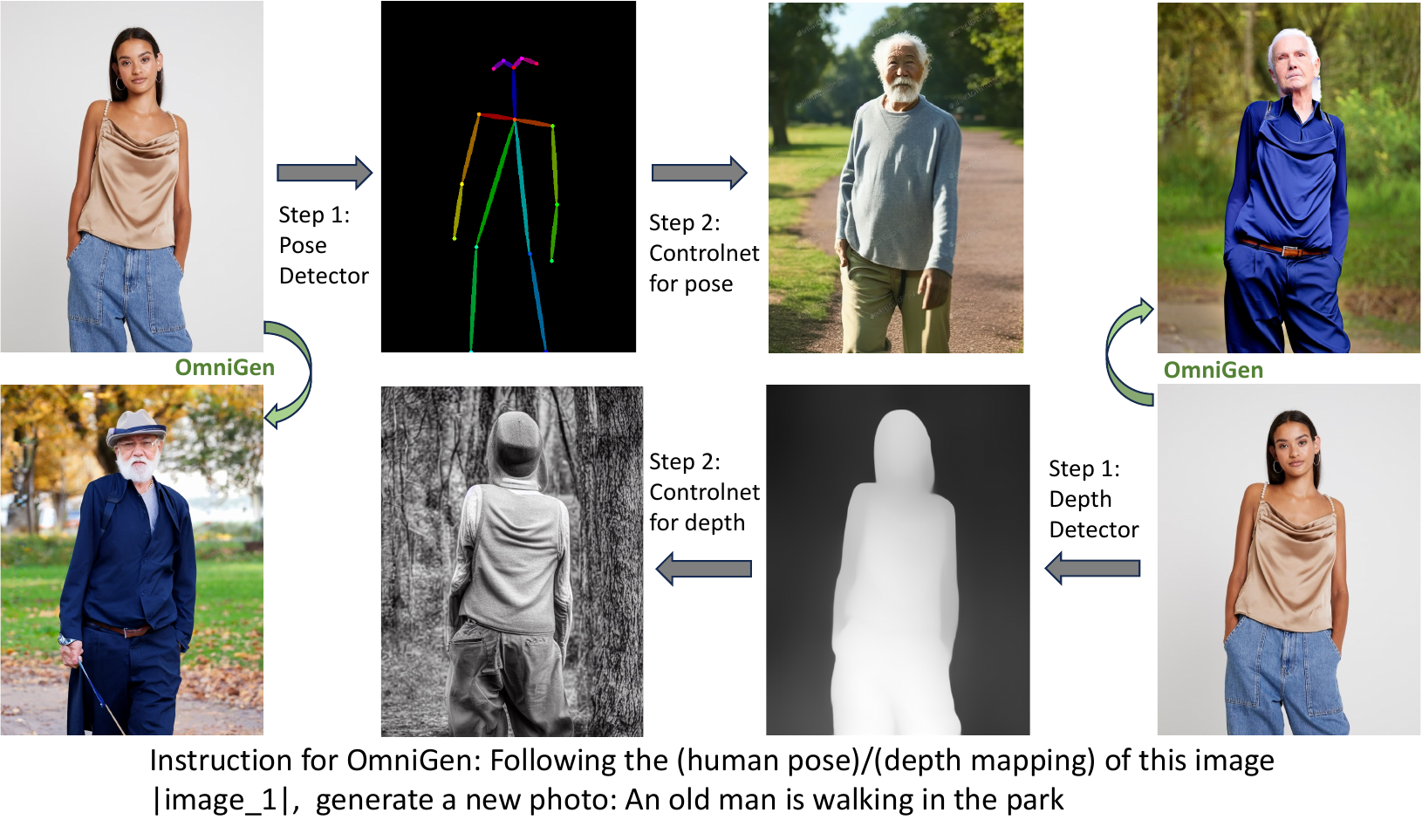}
    \vspace{-0.8cm}
    \caption{ControlNet involves multiple steps and plugins: firstly using the corresponding detector to extract information from the reference image, and then loading the appropriate ControlNet module to process the visual conditions. In contrast,  \model completes the entire task in a single step within a single model.}
    \label{fig:simple_control}
    \vspace{-0.6cm}
\end{figure}

\subsection{Computer Vision Tasks}
\label{sec:vision}

We present several qualitative results of computer vision tasks in Figure~\ref{fig:cv}. 
\model can handle various low-level vision tasks such as deraining, deblurring, and inpainting. 
At the bottom of Figure~\ref{fig:cv}, we can see that \model is also able to handle high-level vision tasks, such as human pose recognition.

Note that OmniGen's capabilities in CV tasks are not intended to surpass existing state-of-the-art models that developed for specific tasks over a long period. 
We hope it can transfer the knowledge gained from these traditional computer vision tasks to image generation tasks, thereby unlocking greater potential.
Some examples are shown in Figure~\ref{fig:simple_control}. 
The existing workflow for ControlNet involves using a detector to extract spatial condition information from the reference image, and then loading the corresponding control module to model the spatial condition information for image generation. 
Can we directly use \model to generate new images based on a reference image in only one step?
Surprisingly, even without having encountered such a task before, \model handles it admirably without explicit extraction of additional conditional images.
More specifically, we can directly input the reference image and text instruction (e.g., \textit{Follow the human pose of this image to generate new image.}) to generate the target image in only one step without any explicit additional intermediate procedures.

\section{Further Analysis}

LLMs demonstrate remarkable generalization capabilities, and can boost performance through mechanisms such as in-context learning and chain of thought. We observe similar functionalities in \model as well, and present our findings in this section.

\subsection{Emgerging Capabilities}

By standardizing all tasks into a unified format and training on \textbf{X2I} dataset,  \model can acquire universal knowledge and allow knowledge transfer across different scenarios and tasks, thus enabling the generation capabilities on unseen tasks and domains.

\begin{figure}
    \centering
    \includegraphics[width=\linewidth]{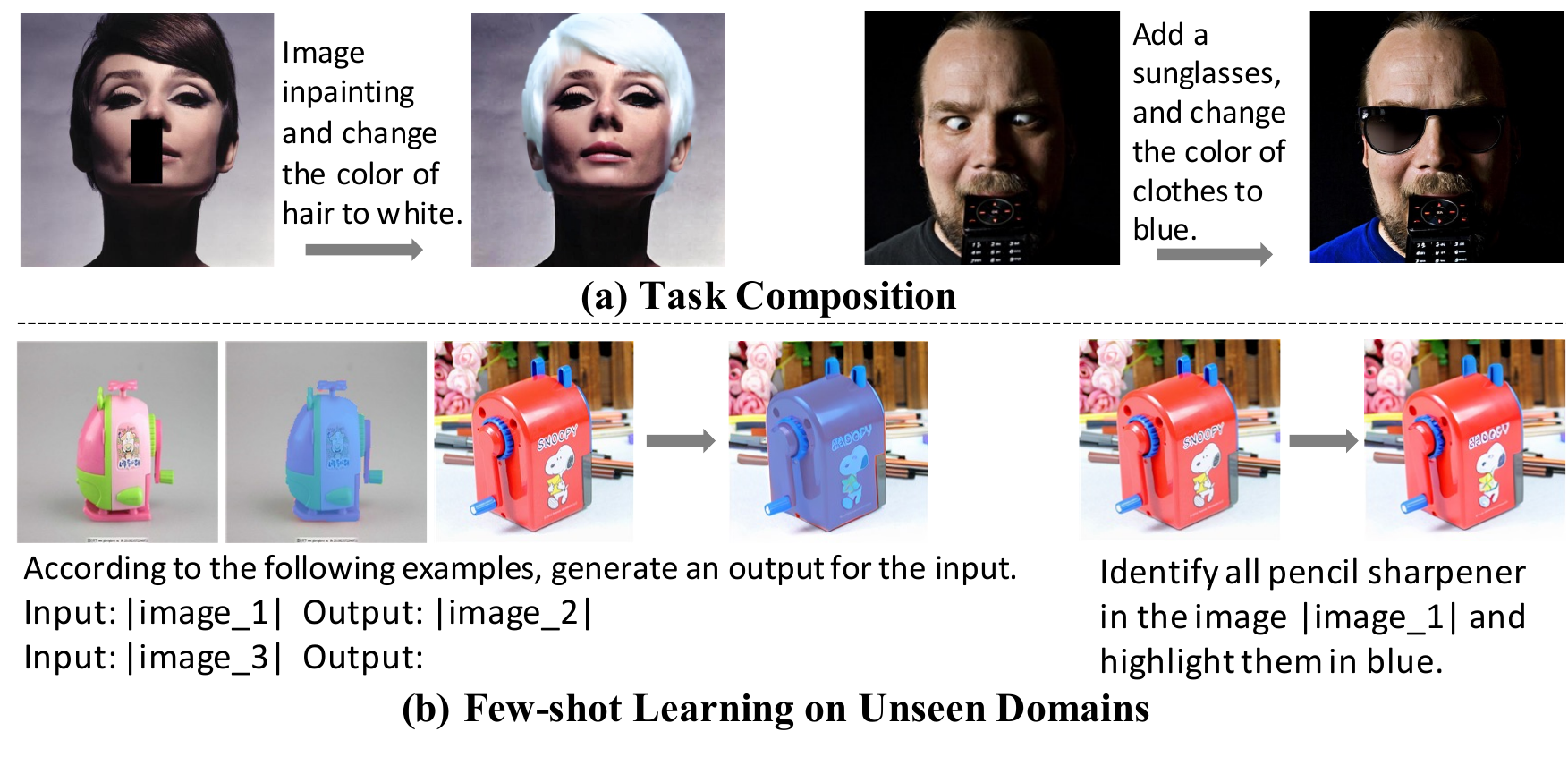}
    \vspace{-0.8cm}
    \caption{Examples of Emgerging Capabilities of \model.}
    \label{fig:emgerging}
    \vspace{-0.6cm}
\end{figure}

\textbf{Task Composition}. 
In real-world applications, user requirements often involve combinations of tasks. As shown in Figure~\ref{fig:emgerging}-(a), our model is capable of simultaneously processing multiple instructions, including those for different tasks as well as multiple instructions for the same task. 
These results highlight our model's versatility and potential for widespread adoption in the wild.

\textbf{End-to-end Workflow}. 
Users typically need to load multiple models and perform multi-step processing to ultimately generate a satisfactory image, making the workflow very cumbersome and costly.
\model possesses both excellent multi-modal understanding and image generation capabilities, enabling it to complete a lot of tasks without relying on external models, thereby significantly simplifying the workflow and saving cost. 
As demonstrated in Figure~\ref{fig:simple_control}, \model can extract the relevant conditional information from the reference image and generate a new image based on the captured condition within one step. This process is implicit, with all processing completed internally within the model, only requiring the user to input a simple command. 

\textbf{In-context Learning for Unseen Domains}. 
We use the data from FSS~\citep{li2020fss} which contains objects that have never been seen in previous datasets to evaluate the generalized ability.
In the right of Figure~\ref{fig:emgerging}-(b), we can see that \model is not familiar with the concepts of ``pencil sharpeners'', and it cannot identify it from images. However, when provided with an example, the model is capable of making accurate predictions. By providing an example, the model is capable of successfully completing this unseen segmentation task.


\subsection{Reasoning Ability}
\begin{figure}
    \centering
    \includegraphics[width=\linewidth]{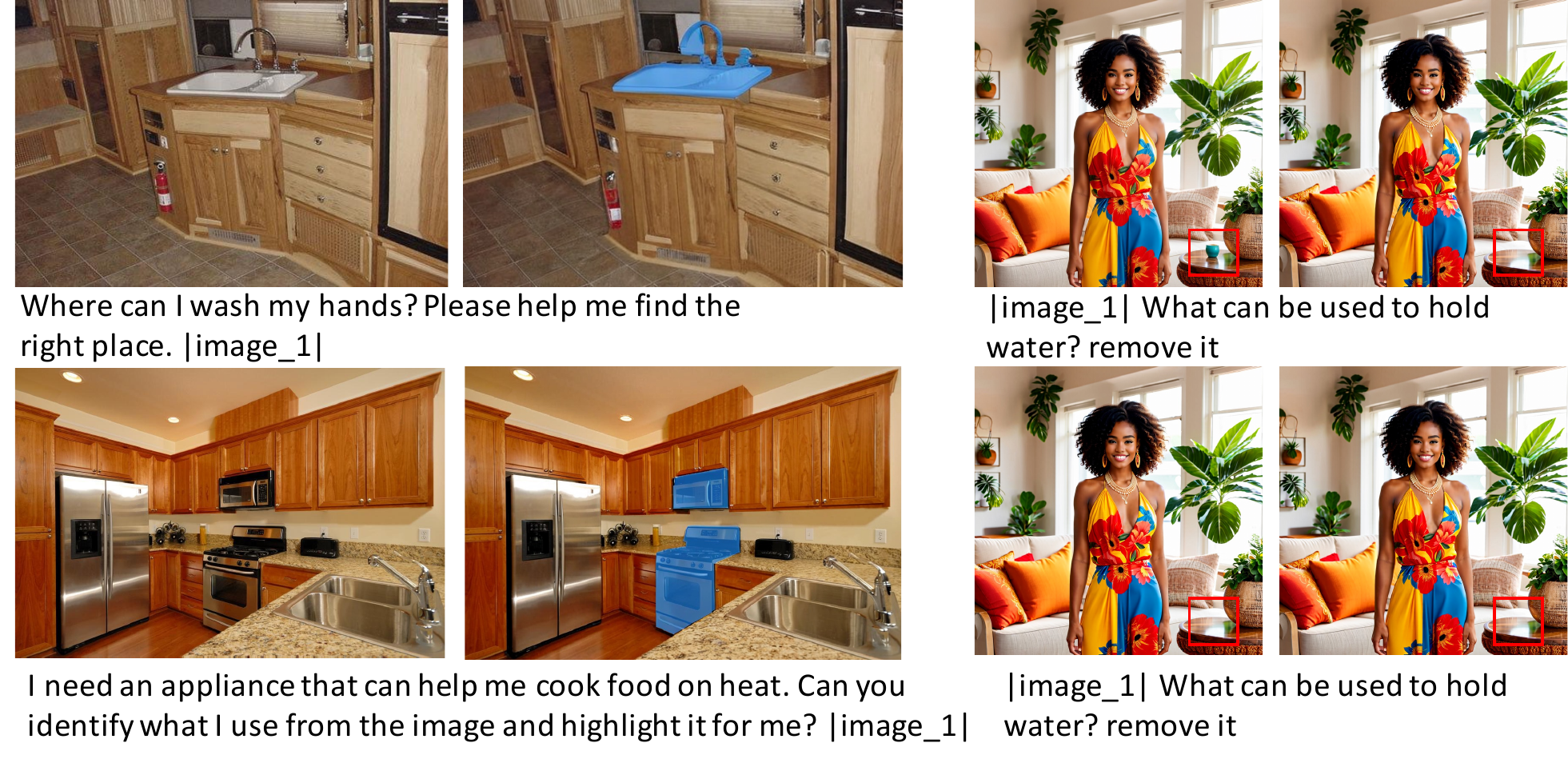}
    \vspace{-0.8cm}
    \caption{Reasoning ability of \model.}
    \vspace{-0.3cm}
    \label{fig:reasoning}
\end{figure}

We have explored the reasoning capabilities of the model and presented the results in Figure~\ref{fig:reasoning}. As shown in the left half of Figure~\ref{fig:reasoning}, when given an instruction without explicitly specifying the object, such as \textit{``Where can I wash my hands? Please help me find the right place in image\_1''}, the model can recognize image contents and infer that a sink is needed. Consequently, the model identifies and indicates the area of the sink in the image. This functionality creates potential applications in the field of embodied intelligence, assisting intelligent agents in comprehending multi-modal instructions, locating necessary objects and planning subsequent actions. Moreover, the right half of Figure~\ref{fig:reasoning} demonstrates that after inferring the target object, the model can also perform editing operations on it. If no object matches, the model will refrain from editing any unrelated objects(see the example in the bottom right corner of Figure~\ref{fig:reasoning}).

\subsection{Chain of Thought}
\begin{figure}
    \centering
    \includegraphics[width=\linewidth]{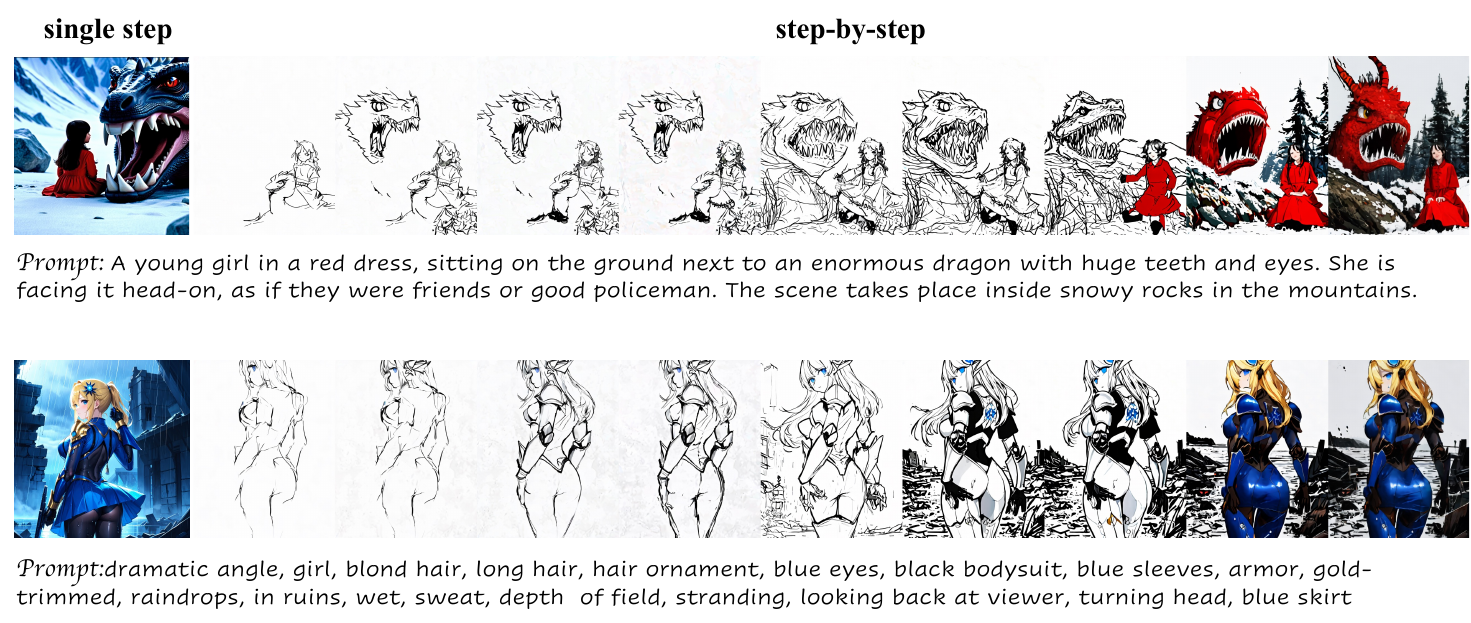}
    \vspace{-0.7cm}
    \caption{Simulate the process of human drawing via step-by-step generation.}
    \vspace{-0.7cm}
    \label{fig:paint-step-by-step}
\end{figure}

The Chain-of-Thought (CoT) method can significantly boost the performance of LLMs by decomposing the task into multiple steps and sequentially solving each step to obtain an accurate final answer. We consider whether a similar alternative can be applied to image generation. Inspired by the basic way of human drawing, we hope to mimic the step-by-step drawing process, iteratively refining the image from a blank canvas. We construct an anime image dataset and use the PAINTS-UNDO\footnote{https://github.com/lllyasviel/Paints-UNDO} model to simulate each stage of artwork creation. We select 8 representative frames to depict the gradual development of the final image. 
After filtering out inconsistent sequences, we fine-tuned the model on this dataset for 16,000 steps. 

The results are visualized in Figure~\ref{fig:paint-step-by-step}, alongside the outputs generated by the original model.
For step-by-step generation, the input data consists of the current step's image and text, and then the model predicts the image for the next step. It can be observed that the fine-tuned model successfully simulates the behavior of a human artist: drawing the basic outline, incrementally adding details, making careful modifications, and applying colors to the image. In this manner, users can modify the previous results to control the current output, thereby participating more actively in the image generation process, rather than passively waiting for the final image with a black-box diffusion model. Unfortunately, the quality of the final generated images does not surpass that of the original model. In the step-by-step generation approach, the model may incorporate erroneous modifications, leading to some disarray in the final image. This does not imply that the approach is unfeasible; currently, we only conduct a preliminary exploration, leaving further optimizations for future research.
Based on the findings of previous work~\citep{lightman2023let} on LLMs, which indicate that process supervision significantly outperforms outcome supervision, we posit that supervising the drawing process of images is a promising direction that may assist the model in handling more complex and diverse scenes.

\section{Ablation Study}

In this section, we assess the necessity of certain modules in OmniGen

\begin{figure}[t]
    \centering
    \includegraphics[width=\linewidth]{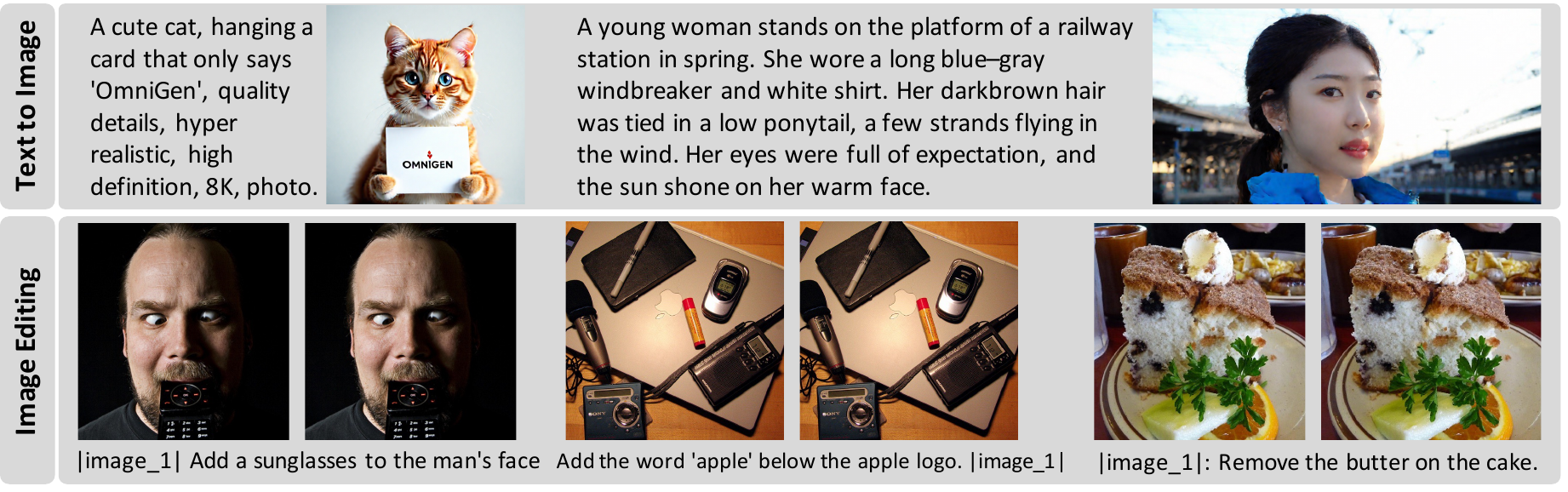}
    \vspace{-0.6cm}
    \caption{\textbf{Top}: results of causal attention; \textbf{Bottom}: results without loss weights.}
    \label{fig:ablation}
    \vspace{-0.3cm}
\end{figure}

\textbf{Attention Mask}. 
Figure~\ref{fig:ablation}-Top shows the generated images without the modified attention in Section~\ref{sec:design}. As we can see, there is a lot of noise in the images. Compared to Figure~\ref{fig:image_gen}, the generated images in Figure~\ref{fig:ablation}-Top have many distorted areas,
confirming that the unidirectional attention mechanism in LLMs is not suitable for the rectified flow.
    
\textbf{Weighted Loss}.
Figure~\ref{fig:ablation}-Bottom shows the results when the weighted loss from Section~\ref{sec:train} is not used for training. 
For image editing tasks, since the edited area is usually small, the model tends to learn to copy the input image directly as the output. Comparing Figure~\ref{fig:ablation}-Bottom and Figure~\ref{fig:image_gen}, it's evident that using weighted loss can prevent the model from learning this shortcut.

\textbf{Input Image Representation}.
The CLIP model is widely used in multi-modal understanding models~\citep{llava}. We also explored the difference between using CLIP\footnote{https://huggingface.co/openai/clip-vit-large-patch14-336}
 and using VAE to process input images. The results are shown in Table~\ref{tab:ablation}. There is not much difference between the two strategies. Using VAE has a slight advantage in image similarity, which might be related to the fact that the output images also require VAE processing. Another advantage of using VAE is that it avoids additional modules, maintaining the simplicity of the model. Directly inputting the original image is another simpler approach that we leave for future exploration.

\begin{table}[t]
\small
\centering
\begin{tabular}{l|cc|cc}
\toprule
 \textbf{Input Image} & \multicolumn{2}{c|}{\textbf{EMU-Edit}} & \multicolumn{2}{c}{\textbf{DreamBench}}  \\
 \textbf{Representation} & CLIP-T & CLIP-I & CLIP-T & CLIP-I \\
\midrule
CLIP & 0.232 & 0.820 & 0.312 & 0.789 \\
VAE & 0.231 & 0.829 & 0.315 & 0.801\\
\bottomrule
\end{tabular}
\vspace{-0.1cm}
\caption{Comparison of different methods to obtain visual embedding for input images.}
\label{tab:ablation}
\vspace{-0.6cm}
\end{table}

%% file: sec/4_related_work.tex
\section{Related Work}

\textbf{Generative Foundation Models}. 
The GPT series~\citep{gpt2,openai2024gpt4technicalreport} have demonstrated that language models can learn numerous tasks via training on a large-scale dataset. 
Beyond language, multi-modal large language models~\citep{llava,chen2024internvl} have been proposed to integrate vision and language capabilities. 
However, they lack the capability to generate images.
Some works propose integrating LLMs with diffusion models to equip LLMs with image generation capability~\citep{sun2024generative,ge2024seed,wu24nextgpt,tang2024codi}. Others use discrete tokens to support both image and text generation simultaneously~\citep{team2024chameleon,lu2024unified,xie2024show}. 
They focus on multi-modal generation with limited image-generation capability.
Concurrent works such as TransFusion~\citep{zhou2024transfusion} and Show-O~\citep{xie2024show} unify diffusion and autoregressive methods into a single model, generating text autoregressively and images through diffusion.  
Nonetheless, like existing diffusion models, these works focus on a limited range of image generation tasks, primarily text-to-image generation, and cannot cover more complex and various visual generation tasks. 
The construction of a universal foundation model for image generation remains unclear and has not been fully explored.

\textbf{Diffusion Model}.
Recent advancements in diffusion models have been remarkable, with notable contributions from the Stable Diffusion (SD) series~\citep{LDM,podell2023sdxl,sd3}, DALL-E~\citep{dalle2}, and Imagen~\citep{imagen3}. These models are predominantly designed for text-to-image generation tasks. 
Many efforts have been made to extend the capabilities of diffusion models, such as ControlNet~\citep{controlnet}, T2I-Adapter~\citep{t2iadapter}, StyleShot~\citep{gao2024styleshot}. 
InstructPix2Pix~\citep{brooks2023instructpix2pix}, InstructSD~\citep{Paul2023instruction-tuning-sd} and EMU-edit~\citep{sheynin2024emuedit} explore performing general image editing tasks through instructions.
However, these methods are task-specific, extending the capabilities of SD by modifying the model architecture. 
In contrast, \model is a model that natively supports various image-generation tasks, and no longer requires any preprocessing steps or assistance from other models.
There is some work exploring the unification of computer vision (CV) tasks~\citep{bachmann20244m,painter,gan2023instructcv,geng2024instructdiffusion}. However, these efforts primarily focus on classic vision tasks instead of general image generation tasks, and often underperform compared to those specifically designed and trained for corresponding tasks. In our work, the introduction of CV tasks does not aim to obtain a SOTA performance on CV tasks, which is designed to enable the model to learn general knowledge.

%% file: sec/5_conclusion.tex
\section{Limitations and Conclusion}

In this work, we introduce OmniGen, the first unified image generation model.
It is designed to be simple, flexible, and easy to use. We also construct the first large-scale unified image generation dataset, X2I, to activate the general capabilities of the OmniGen model. OmniGen has demonstrated excellent image-generation abilities across various tasks.

However, the current model still has some issues. The text rendering capability is limited, and long text can not be accurately generated. The output images may contain undesired details (e.g. abnormal hand). Besides, previously unseen image types (e.g., surface normal map) can hardly be processed as expected. More failure cases are provided in supplementary materials.

We believe that the future paradigm of image generation should be simple and flexible, allowing for the direct generation of various images through any multimodal instructions without complex workflows. OmniGen represents an important step towards a foundational model for universal image generation. We will open-source the relevant resources, hoping to provide insights for the future of image generation.

%% file: sec/X_suppl.tex
\clearpage

\setcounter{page}{1}
\maketitlesupplementary

\section{More Qualitative Results}
In this section, we provide more qualitative results of OmniGen.
Overall, OmniGen achieves competitive or even superior results compared to task-specific open-source models. It's important to note that OmniGen is a general-purpose model and is not specifically trained for any particular task. Additionally, OmniGen has capabilities that many other models lack, allowing it to follow multimodal instructions to complete complex tasks. \textbf{To the best of our knowledge, OmniGen is the first model capable of such flexible control in image generation.}

\subsection{Text to Image}
\begin{figure*}
    \centering
    \includegraphics[width=\linewidth]{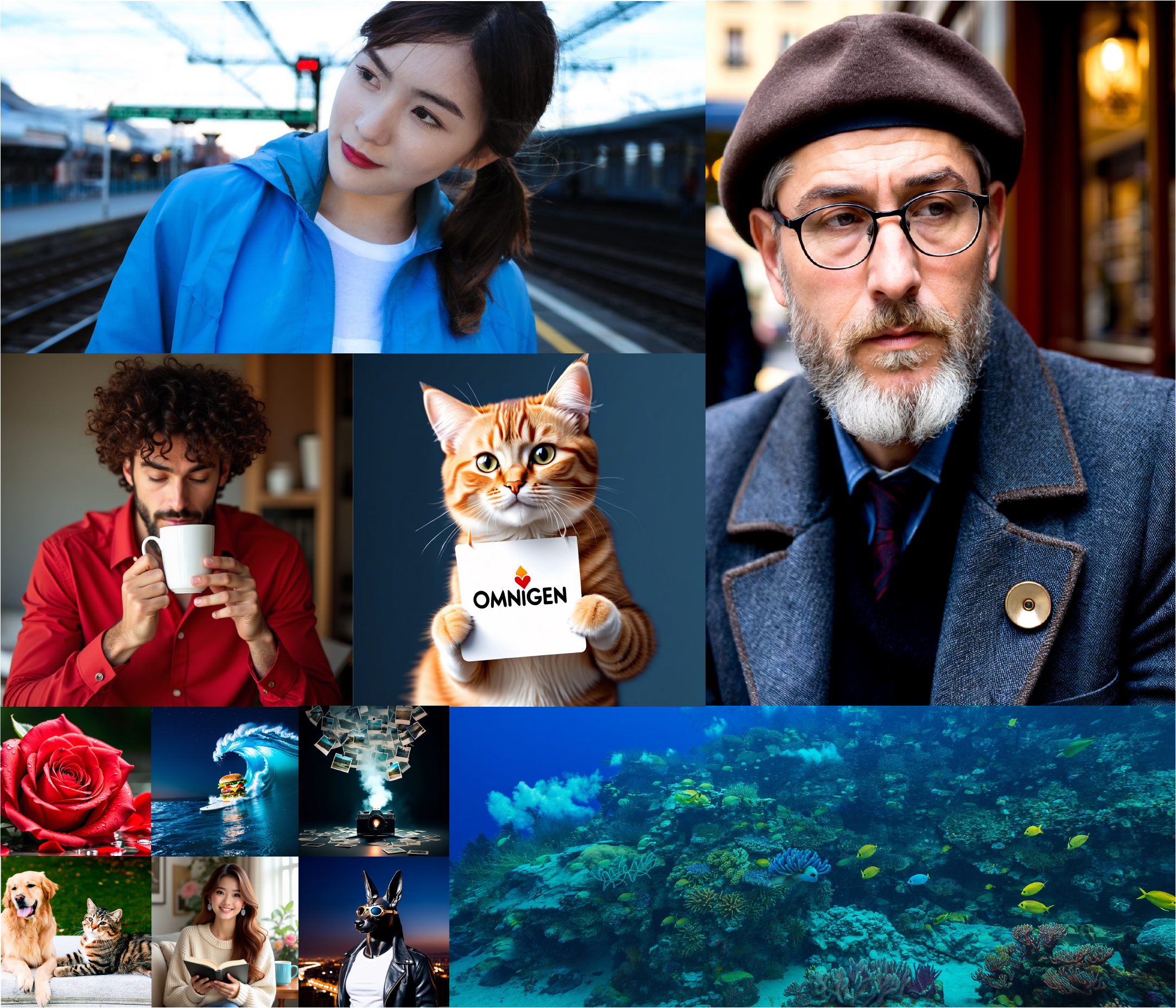}
    \caption{Examples for text-to-image task. OmniGen can generate images with arbitrary aspect ratios.}
    \label{fig:more_t2i}
\end{figure*}

Figure~\ref{fig:more_t2i} shows more examples of text-to-image task.  \model can generate images with arbitrary aspect ratios.

\subsection{Image Editing}
\begin{figure*}
    \centering
    \includegraphics[width=\linewidth]{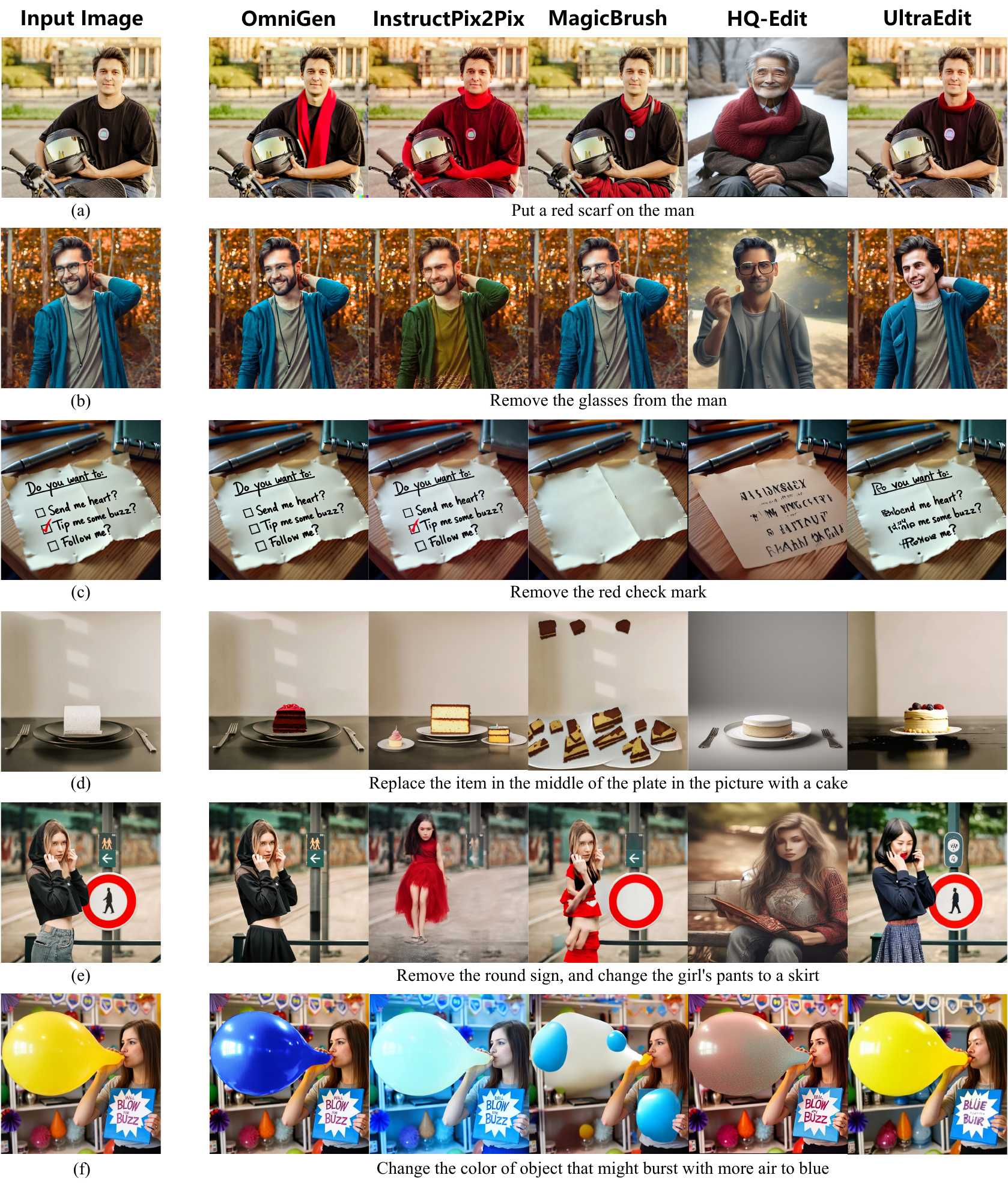}
    \caption{Qualitative results of image editing task. OmniGen achieves better results than other open-source models. Thanks to modeling both text and images within a unified transformer, OmniGen demonstrates strong multimodal instruction understanding and is capable of handling complex tasks, as shown in (f). }
    \label{fig:more_edit}
\end{figure*}

In Figure~\ref{fig:more_edit}, we provide qualitative results of OmniGen compared to some other open-source models. As shown, OmniGen achieves better results than existing open-source models, accurately modifying specified targets without altering other content. Emu-edit is not open-source, so further evaluation was not possible. Additionally, OmniGen demonstrates strong instruction comprehension capabilities, handling complex tasks such as executing two editing instructions in one step, as seen in Figure~\ref{fig:more_edit}-(e), and reasoning editing tasks in the (f). In Figure~\ref{fig:more_edit}-(f), we don't not explicitly specify the target to be modified; the model infers the target (balloon) from the description and makes precise modifications. OmniGen inputs images and text into a unified transformer, allowing for extensive interaction between different modalities, resulting in powerful multimodal instruction following capabilities, which other frameworks like InstructPix2Pix cannot achieve.

\subsection{Subject-driven}
\begin{figure*}
    \centering
    \includegraphics[width=\linewidth]{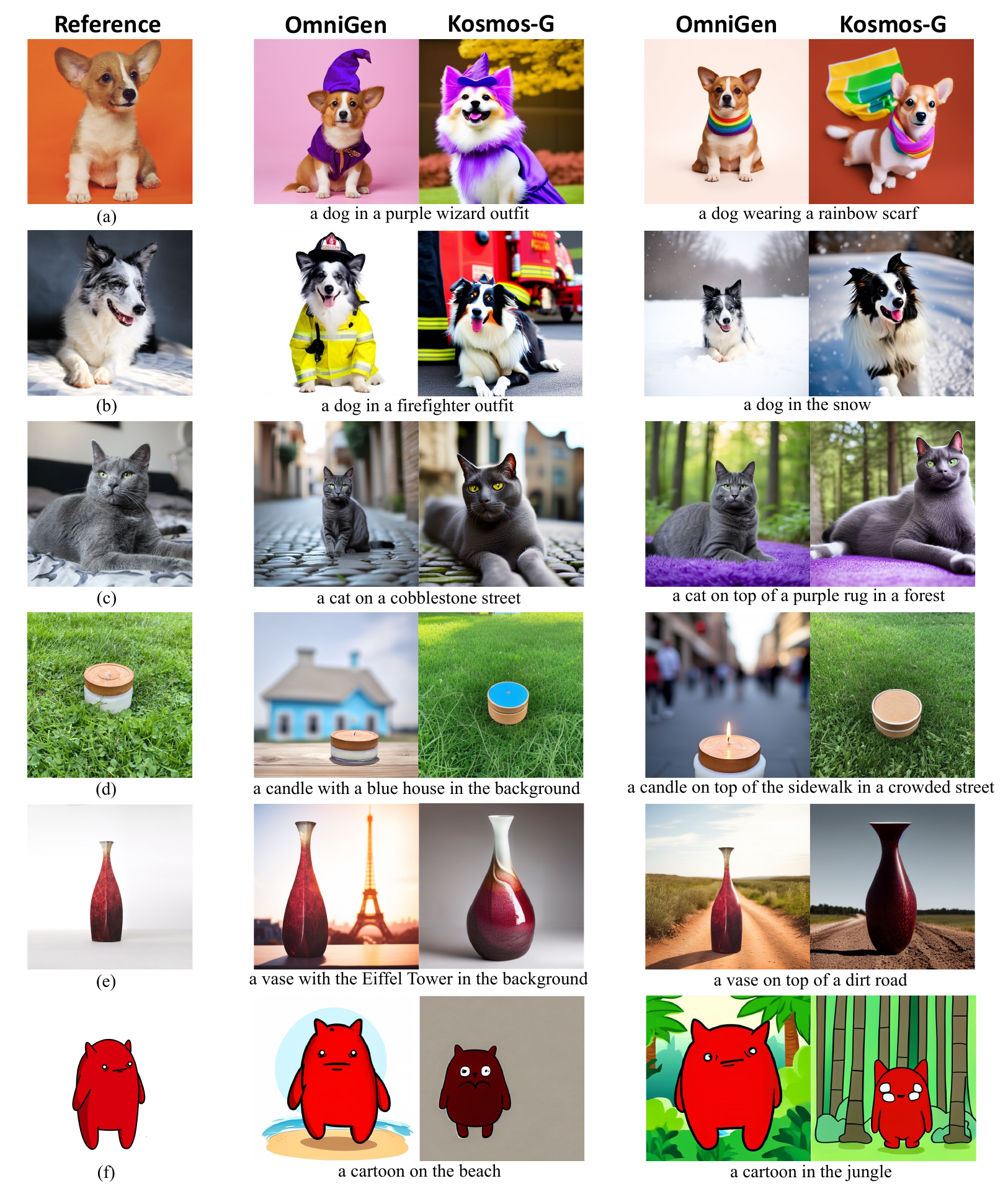}
    \caption{Qualitative results on DreamBench. OmniGen achieves better subject fidelity and text fidelity, preserving the object from the reference image while following new textual instructions. In contrast, Kosmos-G does not always follow text instructions, resulting in a higher CLIP-I similarity. }
    \label{fig:more_subject}
\end{figure*}

Figure~\ref{fig:more_subject} shows more examples of subject-driven generation on DreamBench. Similar to Kosmos-G, we use an image of an object as a reference and generate a new image based on the text description, without fine-tuning on the specific object. As illustrated, OmniGen achieves better subject fidelity and text fidelity, meaning it preserves the object from the reference image while following new textual instructions. Kosmos-G struggles to maintain the object from the reference image. Notably, Kosmos-G does not always adhere to the text instructions; as shown in Figure~\ref{fig:more_subject}-(d), it fails to depict the described content. Instead, Kosmos-G nearly replicates the reference image, which is why it has a higher CLIP-I similarity (see Table 3 in the main paper). Additionally, Kosmos-G uses SD1.5 as the image model, resulting in lower image quality compared to OmniGen.

\subsection{Identity-Preserving Generation}

\begin{figure*}
    \centering
    \includegraphics[width=0.9\linewidth]{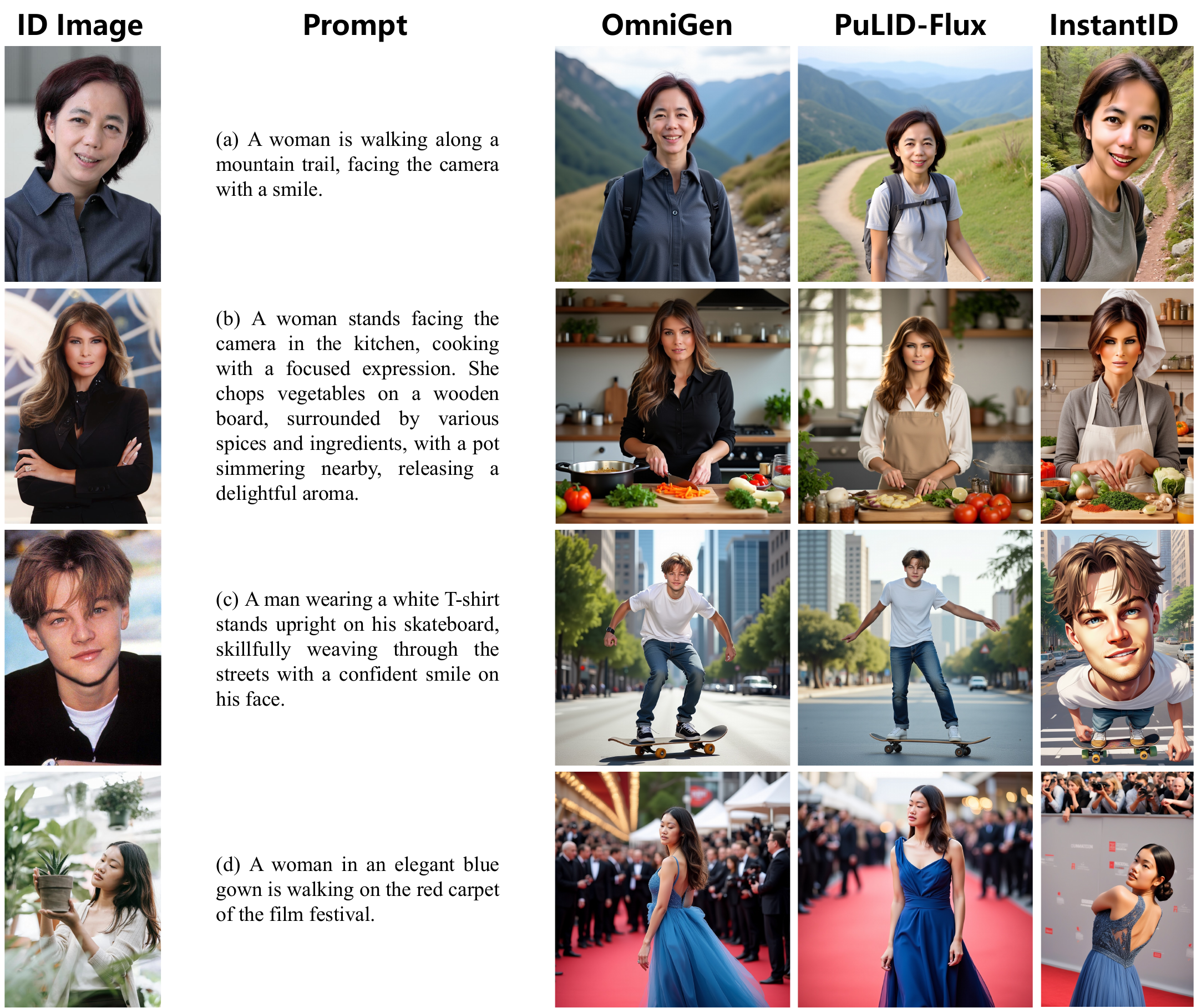}
    \vspace{-0.3cm}
    \caption{Results of Identity-Preserving Generation. OmniGen doesn't require additional face detectors and encoders. OmniGen can utilize clothing information from input images, as shown in examples (a) and (b), and can also follow text instructions to generate new clothing, as seen in examples (c) and (d), demonstrating a higher degree of flexible control.}
    \label{fig:more_id}
    \vspace{-0.4cm}
\end{figure*}

Figure~\ref{fig:more_id} shows the results of Identity-Preserving generation task.
OmniGen achieves better quality than instandID and comparable results to the latest model PULID-FLUX~\footnote{https://huggingface.co/spaces/yanze/PuLID-FLUX}. 
It should be noted that as a general model, OmniGen's facial similarity is not yet on par with the latest specialized model, PULID-FLUX, which is an area for future improvement.
Unlike these models, which require a face detector to detect faces and a face encoder to extract facial features, OmniGen does not need task preprocessing steps or additional plugins. It automatically completes tasks with input images and instructions. Moreover, OmniGen does not only focus on faces; it can capture other useful reference information in the input image, such as clothing. For example, in Figure~\ref{fig:more_id}-(a) and -(b), OmniGen tends to preserve the clothing of the person in the input image. Of course, when clothing is specified in the instructions, OmniGen can also follow the instructions to use new clothing (see Figure~\ref{fig:more_id}-(c) and -(d)). In contrast, other models can only use the extracted face and cannot utilize the original clothing from the input image.

\begin{figure*}
    \centering
    \includegraphics[width=0.9\linewidth]{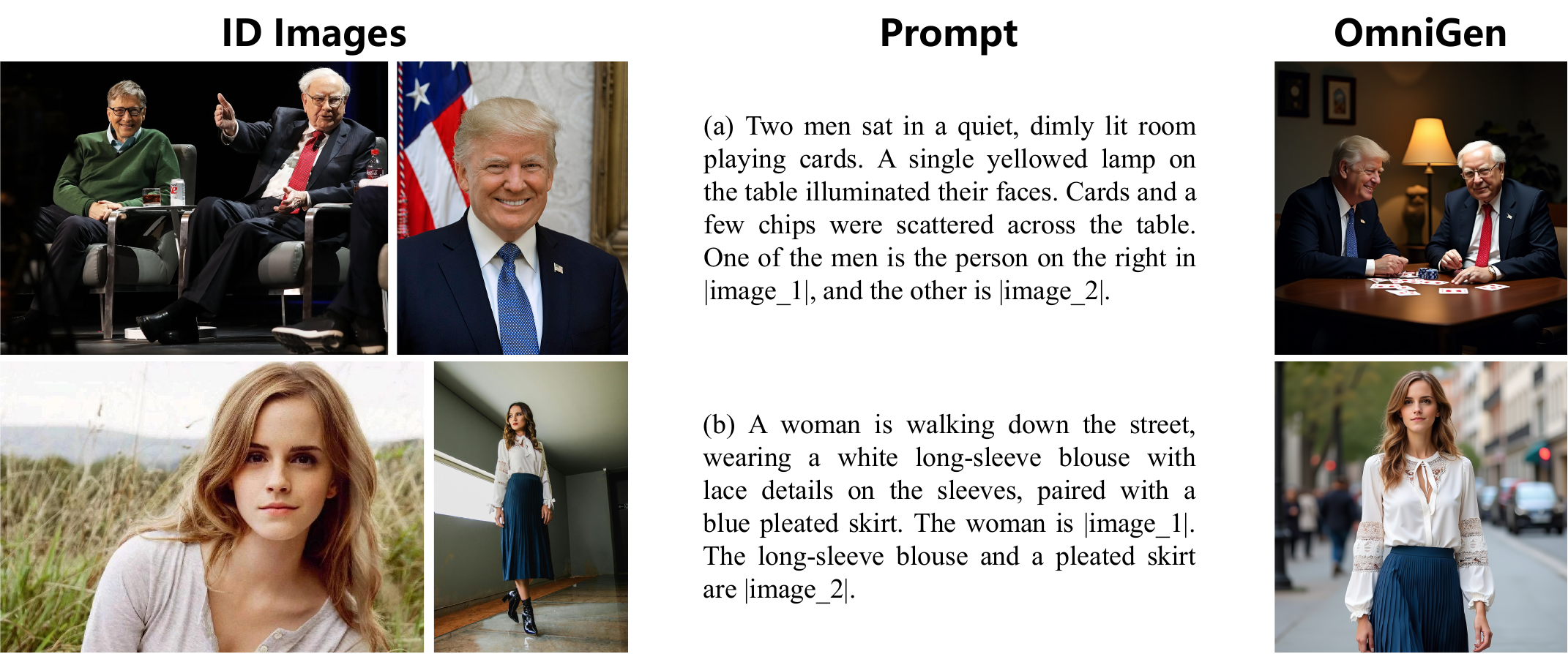}
    \vspace{-0.3cm}
    \caption{Some unique features of OmniGen. OmniGen can not only extract the desired object from image containing multiple objects based on instructions, but also process multiple images simultaneously. To our knowledge, no other model currently can offer this level of flexible and precise control.}
    \label{fig:more_id2}
    \vspace{-0.3cm}
\end{figure*}

Figure~\ref{fig:more_id2} showcases some unique features of OmniGen. Existing models cannot handle input images with multiple objects or multiple input images. OmniGen can extract desired individuals from a group photo based on instructions, and it can also process multiple images simultaneously, as shown in the Figure~\ref{fig:more_id2}-(a). Besides, OmniGen can handle more than just people; as demonstrated in Figure~\ref{fig:more_id2}-(b), it can identify specific clothing items in an image and use them to generate new images. 

\subsection{Failure cases}

\begin{figure*}
    \centering
    \includegraphics[width=\linewidth]{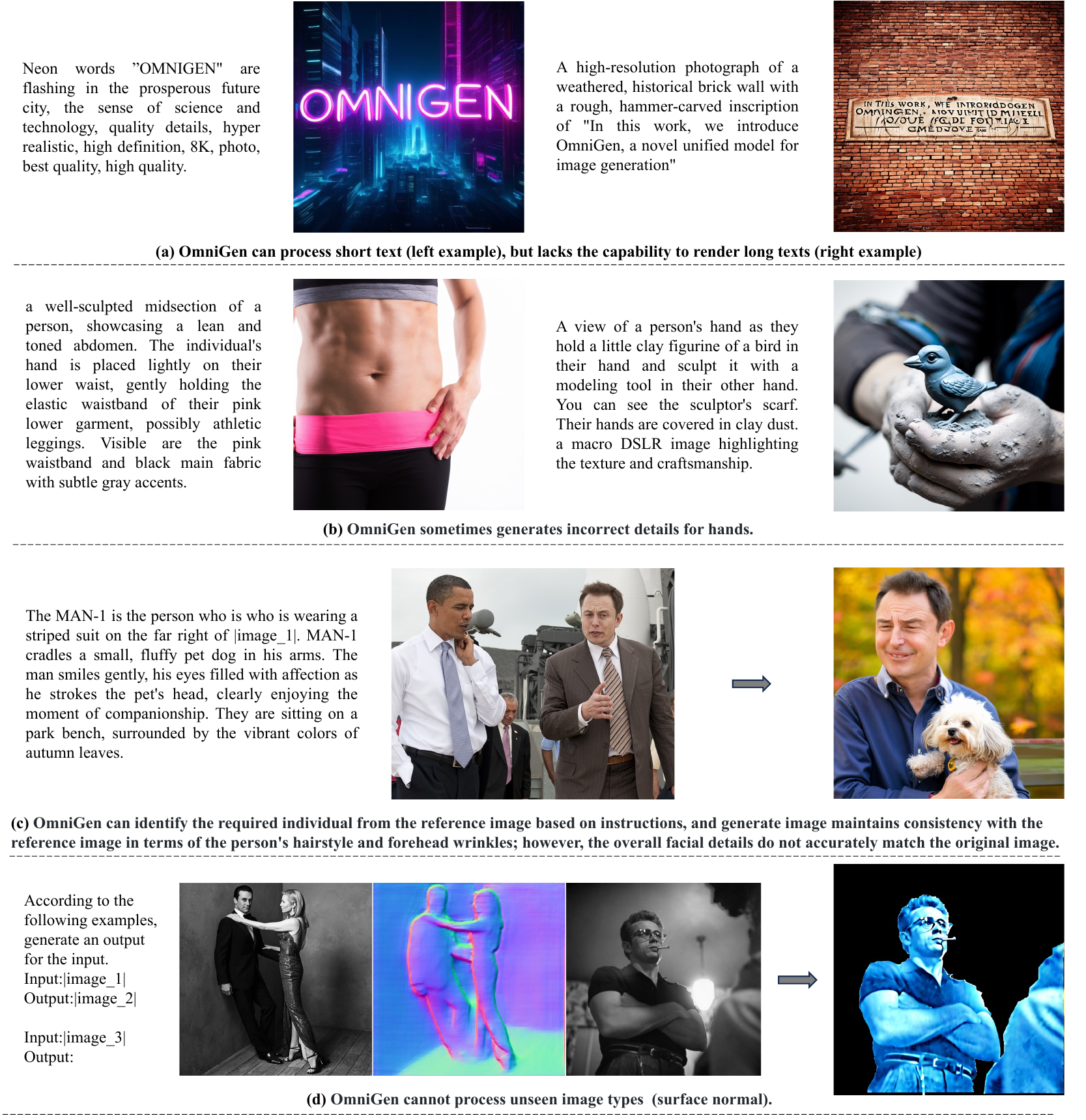}
    \caption{Failure cases of OmniGen}
    \label{fig:failure}
\end{figure*}

Figure~\ref{fig:failure} illustrates several typical failure cases of the current model. We summarize the limitations of the current model as follows:
\begin{itemize}
    \item Similar to existing diffusion models, \model is sensitive to text prompts. Typically, detailed text descriptions result in higher-quality images.
    \item The current model's text rendering capabilities are limited; it can handle short text segments but fails to accurately generate longer texts. Additionally, due to resource constraints, the number of input images during training is limited to a maximum of three, preventing the model from handling long image sequences.  
    \item The generated images may contain erroneous details, especially small and delicate parts. In subject-driven generation tasks, facial features occasionally do not fully align. \model also sometimes generates incorrect depictions of hands.
    \item \model cannot process unseen image types (e.g., image for surface normal estimation).
\end{itemize}

We believe that most limitations can be addressed by training the model on more related data.
Moreover, compared to most models, fine-tuning OmniGen for downstream tasks is simpler.
Due to OmniGen's inherent support for multimodal inputs, users do not need to spend significant effort on designing networks for specific tasks (for example, there is no need to add ControlNet for conditional generation, nor to design additional style encoders for image styles). We will open-source the training code, and users will only need to prepare their data to fine-tune it for various tasks, including new tasks that have never been supported by any model before. We believe this approach can inspire a wide range of interesting downstream applications.

Due to a lack of high-quality image data and insufficient model size, OmniGen's image quality isn't as good as the latest FLUX model. On one hand, expanding the data and model size is crucial; on the other hand, enabling existing models like FLUX to have similar multimodal instruction-following capabilities is also a potential research direction.

\begin{figure*}[t]
    \centering
    \includegraphics[width=0.9\linewidth]{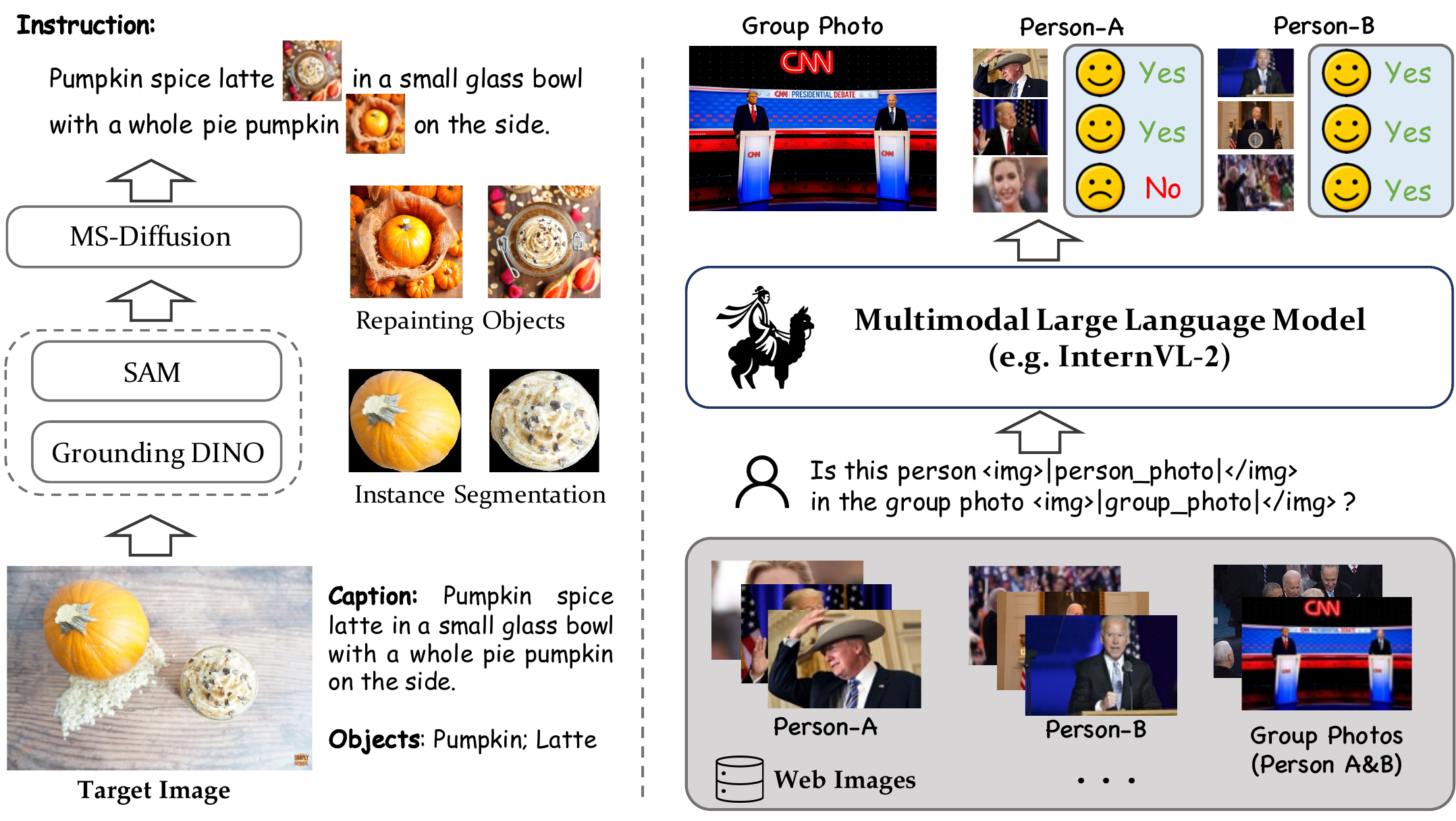}
    \caption{\textbf{Left: Illustration of the construction process for the GRIT-Entity dataset.} We used instance segmentation and repainting methods to acquire a large volume of data. \textbf{Right: Illustration of the cross-verification strategy used in constructing our web images dataset.} For a group photo of Person-A and Person-B, we sampled several images from individual photos of Person-A and Person-B and asked MLLM whether they appear in the group photo. A group photo is retained only if the "Yes" ratio for both Person-A and Person-B meets a specific threshold. The individual images marked as "Yes" are then used to construct data pairs with the corresponding group image.}
    \label{fig:data-cons}
\end{figure*}

\section{Subject-driven dataset}
\label{appendix:data}
Figure~\ref{fig:data-cons}-Left shows the process of constructing the GRIT-Entity dataset. 
Although the GRIT-based approach provides a substantial amount of data, the input data extracted directly from original images can lead the model to fall into simple copy-paste patterns. To fully unleash the subject-driven image generation capability of \model, we constructed a high-quality web images training dataset using natural images of well-known individuals. First, we sampled 20 million Alt-text entries from the Datacomp dataset\footnote{https://huggingface.co/datasets/UCSC-VLAA/Recap-DataComp-1B} and used spaCy\footnote{https://github.com/explosion/spaCy} for named entity recognition. We selected the most frequently occurring names and employed GPT-4o to filter out real, notable individuals, resulting in 2,000 names. Furthermore, we expanded the initial 2,000 names by including closely related individuals, resulting in approximately 10,000 name pairs. We then scraped images of these individuals and pairs from search engines. Due to the noise in web images, where scraped images may not contain the specified individuals, we designed a cross-verification strategy using InternVL~\cite{internvlm2} to filter single and group images, as detailed in~\Cref{fig:data-cons}-Right. The retained single and group images were then captioned with details such as attire and actions. 
We successfully build a dataset of 533K image pairs.


%% file: main.bbl
\begin{thebibliography}{79}
\providecommand{\natexlab}[1]{#1}
\providecommand{\url}[1]{\texttt{#1}}
\expandafter\ifx\csname urlstyle\endcsname\relax
  \providecommand{\doi}[1]{doi: #1}\else
  \providecommand{\doi}{doi: \begingroup \urlstyle{rm}\Url}\fi

\bibitem[Abdin et~al.(2024)Abdin, Jacobs, Awan, Aneja, Awadallah, Awadalla, Bach, Bahree, Bakhtiari, Behl, et~al.]{abdin2024phi3}
Marah Abdin, Sam~Ade Jacobs, Ammar~Ahmad Awan, Jyoti Aneja, Ahmed Awadallah, Hany Awadalla, Nguyen Bach, Amit Bahree, Arash Bakhtiari, Harkirat Behl, et~al.
\newblock Phi-3 technical report: A highly capable language model locally on your phone.
\newblock \emph{arXiv preprint arXiv:2404.14219}, 2024.

\bibitem[Bachmann et~al.(2024)Bachmann, Kar, Mizrahi, Garjani, Gao, Griffiths, Hu, Dehghan, and Zamir]{bachmann20244m}
Roman Bachmann, O{\u{g}}uzhan~Fatih Kar, David Mizrahi, Ali Garjani, Mingfei Gao, David Griffiths, Jiaming Hu, Afshin Dehghan, and Amir Zamir.
\newblock 4m-21: An any-to-any vision model for tens of tasks and modalities.
\newblock \emph{arXiv preprint arXiv:2406.09406}, 2024.

\bibitem[Brooks et~al.(2023)Brooks, Holynski, and Efros]{brooks2023instructpix2pix}
Tim Brooks, Aleksander Holynski, and Alexei~A Efros.
\newblock Instructpix2pix: Learning to follow image editing instructions.
\newblock In \emph{Proceedings of the IEEE/CVF Conference on Computer Vision and Pattern Recognition}, pages 18392--18402, 2023.

\bibitem[Chen et~al.(2024{\natexlab{a}})Chen, Chen, Zhang, Chen, Wu, Zhang, Chen, Li, Wan, and Wang]{chen2024allava}
Guiming~Hardy Chen, Shunian Chen, Ruifei Zhang, Junying Chen, Xiangbo Wu, Zhiyi Zhang, Zhihong Chen, Jianquan Li, Xiang Wan, and Benyou Wang.
\newblock Allava: Harnessing gpt4v-synthesized data for a lite vision-language model, 2024{\natexlab{a}}.

\bibitem[Chen et~al.(2023{\natexlab{a}})Chen, Yu, Ge, Yao, Xie, Wu, Wang, Kwok, Luo, Lu, et~al.]{chen2023pixart}
Junsong Chen, Jincheng Yu, Chongjian Ge, Lewei Yao, Enze Xie, Yue Wu, Zhongdao Wang, James Kwok, Ping Luo, Huchuan Lu, et~al.
\newblock Pixart-alpha: Fast training of diffusion transformer for photorealistic text-to-image synthesis.
\newblock \emph{arXiv preprint arXiv:2310.00426}, 2023{\natexlab{a}}.

\bibitem[Chen et~al.(2023{\natexlab{b}})Chen, Li, Dong, Zhang, He, Wang, Zhao, and Lin]{chen2023sharegpt4v}
Lin Chen, Jisong Li, Xiaoyi Dong, Pan Zhang, Conghui He, Jiaqi Wang, Feng Zhao, and Dahua Lin.
\newblock Sharegpt4v: Improving large multi-modal models with better captions.
\newblock \emph{arXiv preprint arXiv:2311.12793}, 2023{\natexlab{b}}.

\bibitem[Chen et~al.(2022)Chen, Hu, Saharia, and Cohen]{chen2022re_imagen}
Wenhu Chen, Hexiang Hu, Chitwan Saharia, and William~W Cohen.
\newblock Re-imagen: Retrieval-augmented text-to-image generator.
\newblock \emph{arXiv preprint arXiv:2209.14491}, 2022.

\bibitem[Chen et~al.(2024{\natexlab{b}})Chen, Hu, Li, Ruiz, Jia, Chang, and Cohen]{chen2024subject}
Wenhu Chen, Hexiang Hu, Yandong Li, Nataniel Ruiz, Xuhui Jia, Ming-Wei Chang, and William~W Cohen.
\newblock Subject-driven text-to-image generation via apprenticeship learning.
\newblock \emph{Advances in Neural Information Processing Systems}, 36, 2024{\natexlab{b}}.

\bibitem[Chen et~al.(2024{\natexlab{c}})Chen, Hu, Li, Ruiz, Jia, Chang, and Cohen]{suti}
Wenhu Chen, Hexiang Hu, Yandong Li, Nataniel Ruiz, Xuhui Jia, Ming-Wei Chang, and William~W Cohen.
\newblock Subject-driven text-to-image generation via apprenticeship learning.
\newblock \emph{Advances in Neural Information Processing Systems}, 36, 2024{\natexlab{c}}.

\bibitem[Chen et~al.(2024{\natexlab{d}})Chen, Wang, Tian, Ye, Gao, Cui, Tong, Hu, Luo, Ma, et~al.]{internvlm2}
Zhe Chen, Weiyun Wang, Hao Tian, Shenglong Ye, Zhangwei Gao, Erfei Cui, Wenwen Tong, Kongzhi Hu, Jiapeng Luo, Zheng Ma, et~al.
\newblock How far are we to gpt-4v? closing the gap to commercial multimodal models with open-source suites.
\newblock \emph{arXiv preprint arXiv:2404.16821}, 2024{\natexlab{d}}.

\bibitem[Chen et~al.(2024{\natexlab{e}})Chen, Wu, Wang, Su, Chen, Xing, Zhong, Zhang, Zhu, Lu, et~al.]{chen2024internvl}
Zhe Chen, Jiannan Wu, Wenhai Wang, Weijie Su, Guo Chen, Sen Xing, Muyan Zhong, Qinglong Zhang, Xizhou Zhu, Lewei Lu, et~al.
\newblock Internvl: Scaling up vision foundation models and aligning for generic visual-linguistic tasks.
\newblock In \emph{Proceedings of the IEEE/CVF Conference on Computer Vision and Pattern Recognition}, pages 24185--24198, 2024{\natexlab{e}}.

\bibitem[Esser et~al.(2024)Esser, Kulal, Blattmann, Entezari, M{\"u}ller, Saini, Levi, Lorenz, Sauer, Boesel, et~al.]{sd3}
Patrick Esser, Sumith Kulal, Andreas Blattmann, Rahim Entezari, Jonas M{\"u}ller, Harry Saini, Yam Levi, Dominik Lorenz, Axel Sauer, Frederic Boesel, et~al.
\newblock Scaling rectified flow transformers for high-resolution image synthesis.
\newblock In \emph{Forty-first International Conference on Machine Learning}, 2024.

\bibitem[Gal et~al.(2022)Gal, Alaluf, Atzmon, Patashnik, Bermano, Chechik, and Cohen-Or]{text_inv}
Rinon Gal, Yuval Alaluf, Yuval Atzmon, Or Patashnik, Amit~H Bermano, Gal Chechik, and Daniel Cohen-Or.
\newblock An image is worth one word: Personalizing text-to-image generation using textual inversion.
\newblock \emph{arXiv preprint arXiv:2208.01618}, 2022.

\bibitem[Gan et~al.(2023)Gan, Park, Schubert, Philippakis, and Alaa]{gan2023instructcv}
Yulu Gan, Sungwoo Park, Alexander Schubert, Anthony Philippakis, and Ahmed~M Alaa.
\newblock Instructcv: Instruction-tuned text-to-image diffusion models as vision generalists.
\newblock \emph{arXiv preprint arXiv:2310.00390}, 2023.

\bibitem[Gao et~al.(2024{\natexlab{a}})Gao, Liu, Sun, Tang, Zeng, Chen, and Zhao]{gao2024styleshot}
Junyao Gao, Yanchen Liu, Yanan Sun, Yinhao Tang, Yanhong Zeng, Kai Chen, and Cairong Zhao.
\newblock Styleshot: A snapshot on any style.
\newblock \emph{arXiv preprint arXiv:2407.01414}, 2024{\natexlab{a}}.

\bibitem[Gao et~al.(2024{\natexlab{b}})Gao, Zhuo, Lin, Liu, Chen, Du, Xie, Luo, Qiu, Zhang, et~al.]{gao2024lumina}
Peng Gao, Le Zhuo, Ziyi Lin, Chris Liu, Junsong Chen, Ruoyi Du, Enze Xie, Xu Luo, Longtian Qiu, Yuhang Zhang, et~al.
\newblock Lumina-t2x: Transforming text into any modality, resolution, and duration via flow-based large diffusion transformers.
\newblock \emph{arXiv preprint arXiv:2405.05945}, 2024{\natexlab{b}}.

\bibitem[Ge et~al.(2024{\natexlab{a}})Ge, Zhao, Li, Ge, and Shan]{ge2024seededit}
Yuying Ge, Sijie Zhao, Chen Li, Yixiao Ge, and Ying Shan.
\newblock Seed-data-edit technical report: A hybrid dataset for instructional image editing.
\newblock \emph{arXiv preprint arXiv:2405.04007}, 2024{\natexlab{a}}.

\bibitem[Ge et~al.(2024{\natexlab{b}})Ge, Zhao, Zhu, Ge, Yi, Song, Li, Ding, and Shan]{ge2024seed}
Yuying Ge, Sijie Zhao, Jinguo Zhu, Yixiao Ge, Kun Yi, Lin Song, Chen Li, Xiaohan Ding, and Ying Shan.
\newblock Seed-x: Multimodal models with unified multi-granularity comprehension and generation.
\newblock \emph{arXiv preprint arXiv:2404.14396}, 2024{\natexlab{b}}.

\bibitem[Geng et~al.(2024)Geng, Yang, Hang, Li, Gu, Zhang, Bao, Zhang, Li, Hu, et~al.]{geng2024instructdiffusion}
Zigang Geng, Binxin Yang, Tiankai Hang, Chen Li, Shuyang Gu, Ting Zhang, Jianmin Bao, Zheng Zhang, Houqiang Li, Han Hu, et~al.
\newblock Instructdiffusion: A generalist modeling interface for vision tasks.
\newblock In \emph{Proceedings of the IEEE/CVF Conference on Computer Vision and Pattern Recognition}, pages 12709--12720, 2024.

\bibitem[Ghosh et~al.(2024)Ghosh, Hajishirzi, and Schmidt]{ghosh2024geneval}
Dhruba Ghosh, Hannaneh Hajishirzi, and Ludwig Schmidt.
\newblock Geneval: An object-focused framework for evaluating text-to-image alignment.
\newblock \emph{Advances in Neural Information Processing Systems}, 36, 2024.

\bibitem[Goyal et~al.(2017)Goyal, Ebrahimi~Kahou, Michalski, Materzynska, Westphal, Kim, Haenel, Fruend, Yianilos, Mueller-Freitag, et~al.]{goyal2017something}
Raghav Goyal, Samira Ebrahimi~Kahou, Vincent Michalski, Joanna Materzynska, Susanne Westphal, Heuna Kim, Valentin Haenel, Ingo Fruend, Peter Yianilos, Moritz Mueller-Freitag, et~al.
\newblock The" something something" video database for learning and evaluating visual common sense.
\newblock In \emph{Proceedings of the IEEE international conference on computer vision}, pages 5842--5850, 2017.

\bibitem[Guo et~al.(2024)Guo, Wu, Chen, Chen, and He]{guo2024pulid}
Zinan Guo, Yanze Wu, Zhuowei Chen, Lang Chen, and Qian He.
\newblock Pulid: Pure and lightning id customization via contrastive alignment.
\newblock \emph{arXiv preprint arXiv:2404.16022}, 2024.

\bibitem[Han et~al.(2024)Han, Mao, Jiang, Pan, and Zhang]{han2024stylebooth}
Zhen Han, Chaojie Mao, Zeyinzi Jiang, Yulin Pan, and Jingfeng Zhang.
\newblock Stylebooth: Image style editing with multimodal instruction.
\newblock \emph{arXiv preprint arXiv:2404.12154}, 2024.

\bibitem[Ho et~al.(2020)Ho, Jain, and Abbeel]{ddpm}
Jonathan Ho, Ajay Jain, and Pieter Abbeel.
\newblock Denoising diffusion probabilistic models.
\newblock \emph{Advances in neural information processing systems}, 33:\penalty0 6840--6851, 2020.

\bibitem[Imagen-Team-Google(2024)]{imagen3}
Imagen-Team-Google.
\newblock Imagen 3, 2024.

\bibitem[Kazemzadeh et~al.(2014)Kazemzadeh, Ordonez, Matten, and Berg]{refcoco}
Sahar Kazemzadeh, Vicente Ordonez, Mark Matten, and Tamara Berg.
\newblock Referitgame: Referring to objects in photographs of natural scenes.
\newblock In \emph{Proceedings of the 2014 conference on empirical methods in natural language processing (EMNLP)}, pages 787--798, 2014.

\bibitem[Kingma(2013)]{vae}
Diederik~P Kingma.
\newblock Auto-encoding variational bayes.
\newblock \emph{arXiv preprint arXiv:1312.6114}, 2013.

\bibitem[Kirillov et~al.(2023)Kirillov, Mintun, Ravi, Mao, Rolland, Gustafson, Xiao, Whitehead, Berg, Lo, et~al.]{sam}
Alexander Kirillov, Eric Mintun, Nikhila Ravi, Hanzi Mao, Chloe Rolland, Laura Gustafson, Tete Xiao, Spencer Whitehead, Alexander~C Berg, Wan-Yen Lo, et~al.
\newblock Segment anything.
\newblock In \emph{Proceedings of the IEEE/CVF International Conference on Computer Vision}, pages 4015--4026, 2023.

\bibitem[Lai et~al.(2023)Lai, Tian, Chen, Li, Yuan, Liu, and Jia]{lai2023lisa}
Xin Lai, Zhuotao Tian, Yukang Chen, Yanwei Li, Yuhui Yuan, Shu Liu, and Jiaya Jia.
\newblock Lisa: Reasoning segmentation via large language model.
\newblock \emph{arXiv preprint arXiv:2308.00692}, 2023.

\bibitem[Lee et~al.(2022)Lee, Gu, Park, Choi, and Choo]{lee2022hrviton}
Sangyun Lee, Gyojung Gu, Sunghyun Park, Seunghwan Choi, and Jaegul Choo.
\newblock High-resolution virtual try-on with misalignment and occlusion-handled conditions.
\newblock \emph{arXiv preprint arXiv:2206.14180}, 2022.

\bibitem[Li et~al.(2024{\natexlab{a}})Li, Yang, Kuang, Wu, Wang, Xiao, and Chen]{controlnetplus}
Ming Li, Taojiannan Yang, Huafeng Kuang, Jie Wu, Zhaoning Wang, Xuefeng Xiao, and Chen Chen.
\newblock Controlnet++: Improving conditional controls with efficient consistency feedback, 2024{\natexlab{a}}.

\bibitem[Li et~al.(2020)Li, Wei, Chen, Tai, and Tang]{li2020fss}
Xiang Li, Tianhan Wei, Yau~Pun Chen, Yu-Wing Tai, and Chi-Keung Tang.
\newblock Fss-1000: A 1000-class dataset for few-shot segmentation.
\newblock In \emph{Proceedings of the IEEE/CVF conference on computer vision and pattern recognition}, pages 2869--2878, 2020.

\bibitem[Li et~al.(2024{\natexlab{b}})Li, Tu, Hui, Wang, Zhao, Xiao, Ren, Mei, Liu, Zheng, Zhou, and Xie]{li2024recaption}
Xianhang Li, Haoqin Tu, Mude Hui, Zeyu Wang, Bingchen Zhao, Junfei Xiao, Sucheng Ren, Jieru Mei, Qing Liu, Huangjie Zheng, Yuyin Zhou, and Cihang Xie.
\newblock What if we recaption billions of web images with llama-3?
\newblock \emph{arXiv preprint arXiv:2406.08478}, 2024{\natexlab{b}}.

\bibitem[Li et~al.(2024{\natexlab{c}})Li, Zhang, Diao, Wang, Wang, and Duan]{li2024DenseFusion}
Xiaotong Li, Fan Zhang, Haiwen Diao, Yueze Wang, Xinlong Wang, and Ling-Yu Duan.
\newblock Densefusion-1m: Merging vision experts for comprehensive multimodal perception.
\newblock \emph{2407.08303}, 2024{\natexlab{c}}.

\bibitem[Li et~al.(2023)Li, Liu, Wu, Mu, Yang, Gao, Li, and Lee]{li2023gligen}
Yuheng Li, Haotian Liu, Qingyang Wu, Fangzhou Mu, Jianwei Yang, Jianfeng Gao, Chunyuan Li, and Yong~Jae Lee.
\newblock Gligen: Open-set grounded text-to-image generation.
\newblock In \emph{Proceedings of the IEEE/CVF Conference on Computer Vision and Pattern Recognition}, pages 22511--22521, 2023.

\bibitem[Lightman et~al.(2023)Lightman, Kosaraju, Burda, Edwards, Baker, Lee, Leike, Schulman, Sutskever, and Cobbe]{lightman2023let}
Hunter Lightman, Vineet Kosaraju, Yura Burda, Harri Edwards, Bowen Baker, Teddy Lee, Jan Leike, John Schulman, Ilya Sutskever, and Karl Cobbe.
\newblock Let's verify step by step.
\newblock \emph{arXiv preprint arXiv:2305.20050}, 2023.

\bibitem[Liu et~al.(2024)Liu, Li, Wu, and Lee]{llava}
Haotian Liu, Chunyuan Li, Qingyang Wu, and Yong~Jae Lee.
\newblock Visual instruction tuning.
\newblock \emph{Advances in neural information processing systems}, 36, 2024.

\bibitem[Liu et~al.(2023)Liu, Zeng, Ren, Li, Zhang, Yang, Li, Yang, Su, Zhu, et~al.]{groundingdino}
Shilong Liu, Zhaoyang Zeng, Tianhe Ren, Feng Li, Hao Zhang, Jie Yang, Chunyuan Li, Jianwei Yang, Hang Su, Jun Zhu, et~al.
\newblock Grounding dino: Marrying dino with grounded pre-training for open-set object detection.
\newblock \emph{arXiv preprint arXiv:2303.05499}, 2023.

\bibitem[Liu et~al.(2022)Liu, Gong, and Liu]{liu2022flow}
Xingchao Liu, Chengyue Gong, and Qiang Liu.
\newblock Flow straight and fast: Learning to generate and transfer data with rectified flow.
\newblock \emph{arXiv preprint arXiv:2209.03003}, 2022.

\bibitem[Loshchilov(2017)]{adamw}
I Loshchilov.
\newblock Decoupled weight decay regularization.
\newblock \emph{arXiv preprint arXiv:1711.05101}, 2017.

\bibitem[Lu et~al.(2024)Lu, Clark, Lee, Zhang, Khosla, Marten, Hoiem, and Kembhavi]{lu2024unified}
Jiasen Lu, Christopher Clark, Sangho Lee, Zichen Zhang, Savya Khosla, Ryan Marten, Derek Hoiem, and Aniruddha Kembhavi.
\newblock Unified-io 2: Scaling autoregressive multimodal models with vision language audio and action.
\newblock In \emph{Proceedings of the IEEE/CVF Conference on Computer Vision and Pattern Recognition}, pages 26439--26455, 2024.

\bibitem[Mokady et~al.(2023)Mokady, Hertz, Aberman, Pritch, and Cohen-Or]{mokady2023null}
Ron Mokady, Amir Hertz, Kfir Aberman, Yael Pritch, and Daniel Cohen-Or.
\newblock Null-text inversion for editing real images using guided diffusion models.
\newblock In \emph{Proceedings of the IEEE/CVF Conference on Computer Vision and Pattern Recognition}, pages 6038--6047, 2023.

\bibitem[Mou et~al.(2023)Mou, Wang, Xie, Wu, Zhang, Qi, Shan, and Qie]{t2iadapter}
Chong Mou, Xintao Wang, Liangbin Xie, Yanze Wu, Jian Zhang, Zhongang Qi, Ying Shan, and Xiaohu Qie.
\newblock T2i-adapter: Learning adapters to dig out more controllable ability for text-to-image diffusion models, 2023.

\bibitem[Nah et~al.(2017)Nah, Hyun~Kim, and Mu~Lee]{gopro}
Seungjun Nah, Tae Hyun~Kim, and Kyoung Mu~Lee.
\newblock Deep multi-scale convolutional neural network for dynamic scene deblurring.
\newblock In \emph{Proceedings of the IEEE conference on computer vision and pattern recognition}, pages 3883--3891, 2017.

\bibitem[Onoe et~al.(2024)Onoe, Rane, Berger, Bitton, Cho, Garg, Ku, Parekh, Pont-Tuset, Tanzer, et~al.]{onoe2024docci}
Yasumasa Onoe, Sunayana Rane, Zachary Berger, Yonatan Bitton, Jaemin Cho, Roopal Garg, Alexander Ku, Zarana Parekh, Jordi Pont-Tuset, Garrett Tanzer, et~al.
\newblock Docci: Descriptions of connected and contrasting images.
\newblock \emph{arXiv preprint arXiv:2404.19753}, 2024.

\bibitem[OpenAI(2024)]{openai2024gpt4technicalreport}
OpenAI.
\newblock Gpt-4 technical report, 2024.

\bibitem[Pan et~al.(2023)Pan, Dong, Huang, Peng, Chen, and Wei]{pan2023kosmosg}
Xichen Pan, Li Dong, Shaohan Huang, Zhiliang Peng, Wenhu Chen, and Furu Wei.
\newblock Kosmos-g: Generating images in context with multimodal large language models.
\newblock \emph{arXiv preprint arXiv:2310.02992}, 2023.

\bibitem[Paul(2023)]{Paul2023instruction-tuning-sd}
Sayak Paul.
\newblock Instruction-tuning stable diffusion with instructpix2pix.
\newblock \emph{Hugging Face Blog}, 2023.
\newblock https://huggingface.co/blog/instruction-tuning-sd.

\bibitem[Peebles and Xie(2023)]{dit}
William Peebles and Saining Xie.
\newblock Scalable diffusion models with transformers.
\newblock In \emph{Proceedings of the IEEE/CVF International Conference on Computer Vision}, pages 4195--4205, 2023.

\bibitem[Peng et~al.(2023)Peng, Wang, Dong, Hao, Huang, Ma, and Wei]{kosmos-2}
Zhiliang Peng, Wenhui Wang, Li Dong, Yaru Hao, Shaohan Huang, Shuming Ma, and Furu Wei.
\newblock Kosmos-2: Grounding multimodal large language models to the world.
\newblock \emph{arXiv preprint arXiv:2306.14824}, 2023.

\bibitem[Podell et~al.(2023)Podell, English, Lacey, Blattmann, Dockhorn, M{\"u}ller, Penna, and Rombach]{podell2023sdxl}
Dustin Podell, Zion English, Kyle Lacey, Andreas Blattmann, Tim Dockhorn, Jonas M{\"u}ller, Joe Penna, and Robin Rombach.
\newblock Sdxl: Improving latent diffusion models for high-resolution image synthesis.
\newblock \emph{arXiv preprint arXiv:2307.01952}, 2023.

\bibitem[Qin et~al.(2023)Qin, Zhang, Yu, Feng, Yang, Zhou, Wang, Niebles, Xiong, Savarese, et~al.]{qin2023unicontrol}
Can Qin, Shu Zhang, Ning Yu, Yihao Feng, Xinyi Yang, Yingbo Zhou, Huan Wang, Juan~Carlos Niebles, Caiming Xiong, Silvio Savarese, et~al.
\newblock Unicontrol: A unified diffusion model for controllable visual generation in the wild.
\newblock \emph{arXiv preprint arXiv:2305.11147}, 2023.

\bibitem[Radford et~al.(2019)Radford, Wu, Child, Luan, Amodei, Sutskever, et~al.]{gpt2}
Alec Radford, Jeffrey Wu, Rewon Child, David Luan, Dario Amodei, Ilya Sutskever, et~al.
\newblock Language models are unsupervised multitask learners.
\newblock \emph{OpenAI blog}, 1\penalty0 (8):\penalty0 9, 2019.

\bibitem[Ramesh et~al.(2022)Ramesh, Dhariwal, Nichol, Chu, and Chen]{dalle2}
Aditya Ramesh, Prafulla Dhariwal, Alex Nichol, Casey Chu, and Mark Chen.
\newblock Hierarchical text-conditional image generation with clip latents.
\newblock \emph{arXiv preprint arXiv:2204.06125}, 1\penalty0 (2):\penalty0 3, 2022.

\bibitem[Rombach et~al.(2022)Rombach, Blattmann, Lorenz, Esser, and Ommer]{LDM}
Robin Rombach, Andreas Blattmann, Dominik Lorenz, Patrick Esser, and Bj{\"o}rn Ommer.
\newblock High-resolution image synthesis with latent diffusion models.
\newblock In \emph{Proceedings of the IEEE/CVF conference on computer vision and pattern recognition}, pages 10684--10695, 2022.

\bibitem[Ruiz et~al.(2023)Ruiz, Li, Jampani, Pritch, Rubinstein, and Aberman]{ruiz2023dreambooth}
Nataniel Ruiz, Yuanzhen Li, Varun Jampani, Yael Pritch, Michael Rubinstein, and Kfir Aberman.
\newblock Dreambooth: Fine tuning text-to-image diffusion models for subject-driven generation.
\newblock In \emph{Proceedings of the IEEE/CVF conference on computer vision and pattern recognition}, pages 22500--22510, 2023.

\bibitem[Schuhmann et~al.(2022)Schuhmann, Beaumont, Vencu, Gordon, Wightman, Cherti, Coombes, Katta, Mullis, Wortsman, et~al.]{schuhmann2022laion}
Christoph Schuhmann, Romain Beaumont, Richard Vencu, Cade Gordon, Ross Wightman, Mehdi Cherti, Theo Coombes, Aarush Katta, Clayton Mullis, Mitchell Wortsman, et~al.
\newblock Laion-5b: An open large-scale dataset for training next generation image-text models.
\newblock \emph{Advances in Neural Information Processing Systems}, 35:\penalty0 25278--25294, 2022.

\bibitem[Sheynin et~al.(2024)Sheynin, Polyak, Singer, Kirstain, Zohar, Ashual, Parikh, and Taigman]{sheynin2024emuedit}
Shelly Sheynin, Adam Polyak, Uriel Singer, Yuval Kirstain, Amit Zohar, Oron Ashual, Devi Parikh, and Yaniv Taigman.
\newblock Emu edit: Precise image editing via recognition and generation tasks.
\newblock In \emph{Proceedings of the IEEE/CVF Conference on Computer Vision and Pattern Recognition}, pages 8871--8879, 2024.

\bibitem[Sun et~al.(2024{\natexlab{a}})Sun, Pan, Ge, Li, Duan, Wu, Zhang, Zhou, Qin, Wang, et~al.]{sun2024journeydb}
Keqiang Sun, Junting Pan, Yuying Ge, Hao Li, Haodong Duan, Xiaoshi Wu, Renrui Zhang, Aojun Zhou, Zipeng Qin, Yi Wang, et~al.
\newblock Journeydb: A benchmark for generative image understanding.
\newblock \emph{Advances in Neural Information Processing Systems}, 36, 2024{\natexlab{a}}.

\bibitem[Sun et~al.(2024{\natexlab{b}})Sun, Cui, Zhang, Zhang, Yu, Wang, Rao, Liu, Huang, and Wang]{sun2024generative}
Quan Sun, Yufeng Cui, Xiaosong Zhang, Fan Zhang, Qiying Yu, Yueze Wang, Yongming Rao, Jingjing Liu, Tiejun Huang, and Xinlong Wang.
\newblock Generative multimodal models are in-context learners.
\newblock In \emph{Proceedings of the IEEE/CVF Conference on Computer Vision and Pattern Recognition}, pages 14398--14409, 2024{\natexlab{b}}.

\bibitem[Tang et~al.(2024)Tang, Yang, Khademi, Liu, Zhu, and Bansal]{tang2024codi}
Zineng Tang, Ziyi Yang, Mahmoud Khademi, Yang Liu, Chenguang Zhu, and Mohit Bansal.
\newblock Codi-2: In-context interleaved and interactive any-to-any generation.
\newblock In \emph{Proceedings of the IEEE/CVF Conference on Computer Vision and Pattern Recognition}, pages 27425--27434, 2024.

\bibitem[Team(2024)]{team2024chameleon}
Chameleon Team.
\newblock Chameleon: Mixed-modal early-fusion foundation models.
\newblock \emph{arXiv preprint arXiv:2405.09818}, 2024.

\bibitem[Tumanyan et~al.(2023)Tumanyan, Geyer, Bagon, and Dekel]{pnp}
Narek Tumanyan, Michal Geyer, Shai Bagon, and Tali Dekel.
\newblock Plug-and-play diffusion features for text-driven image-to-image translation.
\newblock In \emph{Proceedings of the IEEE/CVF Conference on Computer Vision and Pattern Recognition}, pages 1921--1930, 2023.

\bibitem[Wang et~al.(2024{\natexlab{a}})Wang, Bai, Wang, Qin, and Chen]{wang2024instantid}
Qixun Wang, Xu Bai, Haofan Wang, Zekui Qin, and Anthony Chen.
\newblock Instantid: Zero-shot identity-preserving generation in seconds.
\newblock \emph{arXiv preprint arXiv:2401.07519}, 2024{\natexlab{a}}.

\bibitem[Wang et~al.(2023)Wang, Wang, Cao, Shen, and Huang]{painter}
Xinlong Wang, Wen Wang, Yue Cao, Chunhua Shen, and Tiejun Huang.
\newblock Images speak in images: A generalist painter for in-context visual learning.
\newblock In \emph{Proceedings of the IEEE/CVF Conference on Computer Vision and Pattern Recognition}, pages 6830--6839, 2023.

\bibitem[Wang et~al.(2024{\natexlab{b}})Wang, Fu, Huang, He, and Jiang]{ms-diffusion}
X Wang, Siming Fu, Qihan Huang, Wanggui He, and Hao Jiang.
\newblock Ms-diffusion: Multi-subject zero-shot image personalization with layout guidance.
\newblock \emph{arXiv preprint arXiv:2406.07209}, 2024{\natexlab{b}}.

\bibitem[Wei et~al.(2018)Wei, Wang, Yang, and Liu]{lightenhance}
Chen Wei, Wenjing Wang, Wenhan Yang, and Jiaying Liu.
\newblock Deep retinex decomposition for low-light enhancement.
\newblock \emph{arXiv preprint arXiv:1808.04560}, 2018.

\bibitem[Wei et~al.(2023)Wei, Zhang, Ji, Bai, Zhang, and Zuo]{wei2023elite}
Yuxiang Wei, Yabo Zhang, Zhilong Ji, Jinfeng Bai, Lei Zhang, and Wangmeng Zuo.
\newblock Elite: Encoding visual concepts into textual embeddings for customized text-to-image generation.
\newblock In \emph{Proceedings of the IEEE/CVF International Conference on Computer Vision}, pages 15943--15953, 2023.

\bibitem[Wu et~al.(2024)Wu, Fei, Qu, Ji, and Chua]{wu24nextgpt}
Shengqiong Wu, Hao Fei, Leigang Qu, Wei Ji, and Tat-Seng Chua.
\newblock {NE}x{T}-{GPT}: Any-to-any multimodal {LLM}.
\newblock In \emph{Proceedings of the International Conference on Machine Learning}, pages 53366--53397, 2024.

\bibitem[Xiao et~al.(2023)Xiao, Yin, Freeman, Durand, and Han]{xiao2023fastcomposer}
Guangxuan Xiao, Tianwei Yin, William~T Freeman, Fr{\'e}do Durand, and Song Han.
\newblock Fastcomposer: Tuning-free multi-subject image generation with localized attention.
\newblock \emph{arXiv preprint arXiv:2305.10431}, 2023.

\bibitem[Xie et~al.(2024)Xie, Mao, Bai, Zhang, Wang, Lin, Gu, Chen, Yang, and Shou]{xie2024show}
Jinheng Xie, Weijia Mao, Zechen Bai, David~Junhao Zhang, Weihao Wang, Kevin~Qinghong Lin, Yuchao Gu, Zhijie Chen, Zhenheng Yang, and Mike~Zheng Shou.
\newblock Show-o: One single transformer to unify multimodal understanding and generation.
\newblock \emph{arXiv preprint arXiv:2408.12528}, 2024.

\bibitem[Ye et~al.(2023)Ye, Zhang, Liu, Han, and Yang]{ye2023ipadapter}
Hu Ye, Jun Zhang, Sibo Liu, Xiao Han, and Wei Yang.
\newblock Ip-adapter: Text compatible image prompt adapter for text-to-image diffusion models.
\newblock \emph{arXiv preprint arXiv:2308.06721}, 2023.

\bibitem[Zamir et~al.(2022)Zamir, Arora, Khan, Hayat, Khan, Yang, and Shao]{derain}
Syed~Waqas Zamir, Aditya Arora, Salman Khan, Munawar Hayat, Fahad~Shahbaz Khan, Ming-Hsuan Yang, and Ling Shao.
\newblock Learning enriched features for fast image restoration and enhancement.
\newblock \emph{IEEE transactions on pattern analysis and machine intelligence}, 45\penalty0 (2):\penalty0 1934--1948, 2022.

\bibitem[Zhang et~al.(2024)Zhang, Mo, Chen, Sun, and Su]{zhang2024magicbrush}
Kai Zhang, Lingbo Mo, Wenhu Chen, Huan Sun, and Yu Su.
\newblock Magicbrush: A manually annotated dataset for instruction-guided image editing.
\newblock \emph{Advances in Neural Information Processing Systems}, 36, 2024.

\bibitem[Zhang et~al.(2023)Zhang, Rao, and Agrawala]{controlnet}
Lvmin Zhang, Anyi Rao, and Maneesh Agrawala.
\newblock Adding conditional control to text-to-image diffusion models.
\newblock In \emph{Proceedings of the IEEE/CVF International Conference on Computer Vision (ICCV)}, pages 3836--3847, 2023.

\bibitem[Zhao et~al.(2024)Zhao, Chen, Chen, Bao, Hao, Yuan, and Wong]{zhao2024uni}
Shihao Zhao, Dongdong Chen, Yen-Chun Chen, Jianmin Bao, Shaozhe Hao, Lu Yuan, and Kwan-Yee~K Wong.
\newblock Uni-controlnet: All-in-one control to text-to-image diffusion models.
\newblock \emph{Advances in Neural Information Processing Systems}, 36, 2024.

\bibitem[Zheng et~al.(2019)Zheng, Song, Chen, Hu, Cao, and Nie]{fashiontryon}
Na Zheng, Xuemeng Song, Zhaozheng Chen, Linmei Hu, Da Cao, and Liqiang Nie.
\newblock Virtually trying on new clothing with arbitrary poses.
\newblock In \emph{Proceedings of the 27th ACM international conference on multimedia}, pages 266--274, 2019.

\bibitem[Zhou et~al.(2019)Zhou, Zhao, Puig, Xiao, Fidler, Barriuso, and Torralba]{ade20k}
Bolei Zhou, Hang Zhao, Xavier Puig, Tete Xiao, Sanja Fidler, Adela Barriuso, and Antonio Torralba.
\newblock Semantic understanding of scenes through the ade20k dataset.
\newblock \emph{International Journal of Computer Vision}, 127:\penalty0 302--321, 2019.

\bibitem[Zhou et~al.(2024)Zhou, Yu, Babu, Tirumala, Yasunaga, Shamis, Kahn, Ma, Zettlemoyer, and Levy]{zhou2024transfusion}
Chunting Zhou, Lili Yu, Arun Babu, Kushal Tirumala, Michihiro Yasunaga, Leonid Shamis, Jacob Kahn, Xuezhe Ma, Luke Zettlemoyer, and Omer Levy.
\newblock Transfusion: Predict the next token and diffuse images with one multi-modal model.
\newblock \emph{arXiv preprint arXiv:2408.11039}, 2024.

\end{thebibliography}
